\documentclass[12pt,preprint,sort&compress]{elsarticle}

\journal{MSSP}

\usepackage[english]{babel}

\usepackage{microtype}

\usepackage{amssymb, bm, amsmath}

\let\argmin\relax
\let\argmax\relax
\let\diag\relax
\let\d\relax
\DeclareMathOperator*{\argmin}{argmin}
\DeclareMathOperator*{\argmax}{argmax}
\DeclareMathOperator{\diag}{diag}
\DeclareMathOperator{\d}{d} 

\renewcommand{\bm}[1]{\boldsymbol{\mathbf{#1}}} 

\usepackage{natbib}

\usepackage{xcolor}
\usepackage[hidelinks,colorlinks]{hyperref}
\AtBeginDocument{%
\hypersetup{
  urlcolor     = magenta, 
  linkcolor    = violet, 
  citecolor   = teal 
}}

\usepackage[noabbrev]{cleveref}

\usepackage{amssymb, bm, amsmath}

\let\argmin\relax
\let\argmax\relax
\let\diag\relax
\let\d\relax
\DeclareMathOperator*{\argmin}{argmin}
\DeclareMathOperator*{\argmax}{argmax}
\DeclareMathOperator{\diag}{diag}
\DeclareMathOperator{\d}{d} 

\renewcommand{\bm}[1]{\boldsymbol{\mathbf{#1}}} 

\begin{document}

\begin{frontmatter}

    \title{\Large Bayesian Modelling of Multivalued Power Curves from an Operational Wind Farm}

    \author[add1]{L.A.\ Bull \corref{cor1}}
    \ead{l.a.bull@sheffield.ac.uk}
    \author[add1]{P.A.\ Gardner}
    \author[add1]{T.J.\ Rogers}
    \author[add1]{N.\ Dervilis}
    \author[add1]{E.J.\ Cross}
    \author[add2]{E.~Papatheou}
    \author[add3]{A.E.\ Maguire}
    \author[add3]{C.\ Campos}
    \author[add1]{K.\ Worden}

    \cortext[cor1]{Corresponding author}
    
    \address[add1]{Dynamics Research Group, Department of Mechanical Engineering, University of Sheffield, Mappin Street, Sheffield S1 3JD, UK} 
    \address[add2]{College of Engineering Mathematics and Physical Sciences, University of Exeter, Exeter EX4 4QF, UK}
    \address[add3]{Vattenfall Research and Development, Holyrood Road, Edinburgh EH8 8AE, UK}

    \begin{abstract}
        Power curves capture the relationship between wind speed and output power for a specific wind turbine. %
        Accurate regression models of this function prove useful in monitoring, maintenance, design, and planning. %
        In practice, however, the measurements do not always correspond to the ideal curve:  power curtailments will appear as (additional) functional components. %
        Such multivalued relationships cannot be modelled by conventional regression, and the associated data are usually removed during pre-processing. %
        The current work suggests an alternative method to infer multivalued relationships in curtailed power data. %
        Using a population-based approach, an overlapping mixture of probabilistic regression models is applied to signals recorded from turbines within an operational wind farm. %
        The model is shown to provide an accurate representation of practical power data across the population. %
    \end{abstract}

    \begin{keyword}
        \small
            Wind energy; Power curve; Performance monitoring; Structural health monitoring; Gaussian processes; Power Curtailment
    \end{keyword}

\end{frontmatter}


\section{Introduction}
Given an increased demand for renewable energy, accurate predictive models are essential to justify, manage, and monitor wind turbine power generation. In particular, accurate predictions of the \emph{power output} (under uncertainty) enable reliable forecasting of the expected income for a complete wind farm -- as well as individual turbines -- to support the expansion of wind-based energy \cite{TASLIMIRENANI2016544}. %
Robust models of the power output have potential applications in {performance monitoring} and operator control, to ensure optimal use \textit{in situ} \cite{papatheou2017performance,YANG2013365}.

\begin{figure}[pt]
\centering
\includegraphics[width=\linewidth]{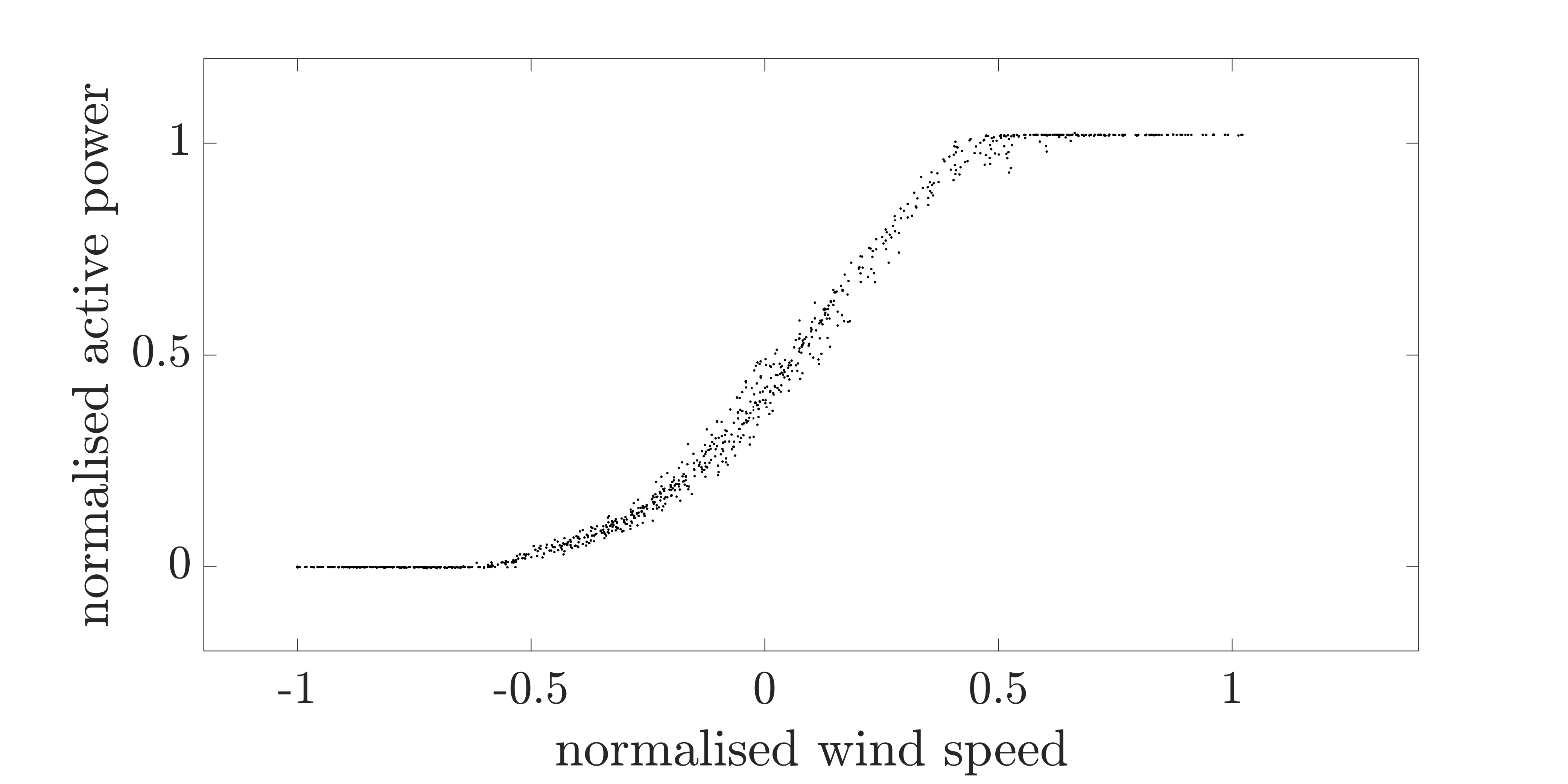}
\caption{Data that represent an ideal power curve. Measurements from three turbines over a period of three weeks.}\label{fig:ideal_data}
\end{figure}

\emph{Power curves} capture the relationship between wind speed and turbine output power \cite{papatheou2017performance} -- the associated function can be used as a key indicator of performance \cite{ROGERS20201124}. %
A regression can be inferred to approximate the relationship given operational measurements (training data) -- typically recorded using Supervisor Control and Sensory Data Acquisition (SCADA) systems \cite{YANG2013365}. %
An example of SCADA data is presented in \Cref{fig:ideal_data}; a regression of these data should generalise to future measurements given \emph{optimal} operation of the wind turbine. %

Various techniques have been proposed to model training data that correspond to \emph{ideal} operation~\cite{thapar2011critical,carrillo2013review,lydia2014comprehensive}. %
In practice, however, only a subset of measurements will typically represent this relationship. %
In particular, power \textit{curtailments} will appear as additional functional components; %
these usually correspond to the output power being controlled (or limited) by the operator. Reasons to limit power include: requirements of the electrical grid \cite{waite2016modeling,hur2014curtailment}, the mitigation of loading/wake effects~\cite{bontekoning2017analysis}, and restrictions enforced by planning regulations. %

\begin{figure}[pt]
\centering
\includegraphics[width=\linewidth]{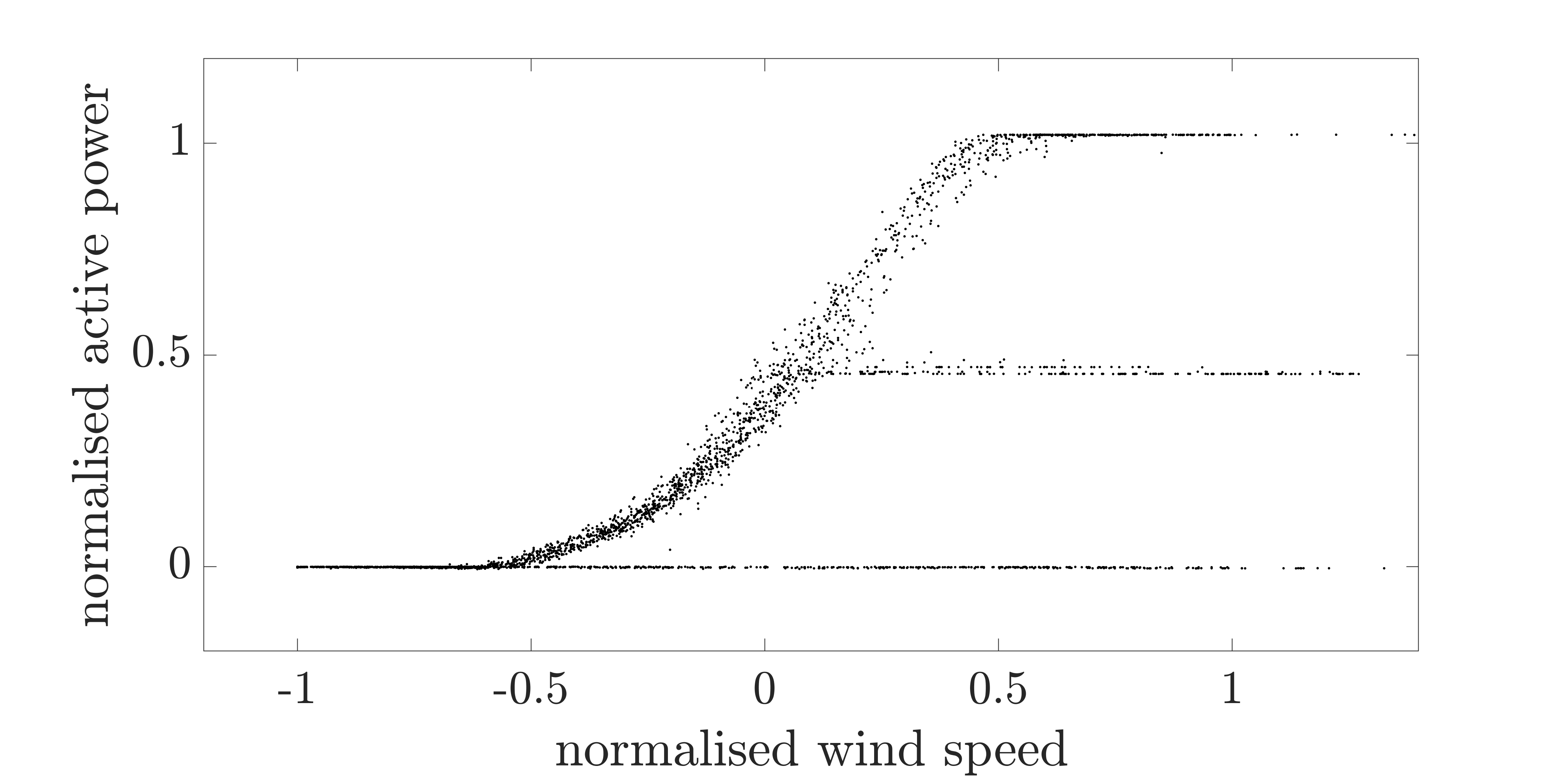}
\caption{Data including power curtailments -- corresponding to (a) the ideal power curve (b) $\approx$50\%-limited output, and (c) zero-limited output. Measurements from seven turbines over nine weeks.}\label{fig:curtailed_data}
\end{figure}

An example of operational data including curtailments is shown in~\Cref{fig:curtailed_data}. %
The emergent space is multivalued, differing significantly from the archetypal curve in~\Cref{fig:ideal_data} (additionally, it cannot be modelled by conventional regression). %
Typically, the curtailment data are removed during pre-processing via engineering judgement \cite{papatheou2017performance}, alongside filtering \cite{MANOBEL20181015, marvuglia2012monitoring} and outlier analysis \cite{MARCIUKAITIS2017732,Papatheou2015} (see~\Cref{s:curts} for details). %

Disregarding curtailment data is logical when modelling the \emph{ideal} curve, corresponding to optimal operation \cite{ROGERS20201124}; despite this fact, curtailed observations are expected in practice. 
Therefore, a representation of \emph{in situ} measurements should model these data, rather than filtering them out, particularly in monitoring or forecasting applications (outlined in~\Cref{s:curts}). %

The current work suggests an overlapping mixture of probabilistic regression models \cite{lazaro2012overlapping} (i.e.\ Gaussian processes \cite{rasmussenGP}) to infer multivalued power curves -- such as those in Figure~\ref{fig:curtailed_data}. %
The statistical method can represent operational power data, including curtailments, while negating requirements for user annotation of the observed data -- i.e.\ categorisation of curtailments is unsupervised. %
As a result, the model can represent observations that might be recorded from \textit{in-situ} turbines in operation (rather than the ideal case only), without the need for extensive outlier analysis, filtering, or pre-processing. %

\section{Related Work}
This work relates to existing literature (e.g.\ \cite{papatheou2017performance,Papatheou2015,YANG2013365}) concerning performance monitoring and prediction via wind turbine power curves. %
As aforementioned, numerous data-based models have been investigated, many of which have been summarised in review papers \cite{thapar2011critical,carrillo2013review,lydia2014comprehensive}. %
A brief summary is provided. 

\emph{Parametric} methods fit parametrised functions to power curve data; some examples include polynomials and sigmoid (tanh/logistic) functions \cite{lydia2014comprehensive,MARCIUKAITIS2017732,TASLIMIRENANI2016544}. Parametric models are desirable -- sigmoid-type functions in particular -- as properties that appear inherent to power curves can be included; for example: the cut-in/cut-out wind speeds, bounded power above and below these values, as well as \emph{near-linear} behaviour within the bounds. %
Unfortunately, over-simplified functions can prove restrictive when approximating the wind-power relationship, while overly complex models (e.g.\ high-order polynomials) are susceptible to overtraining, and require validation procedures to ensure good generalisation to new data \cite{ROGERS20201124}. %

Alternative methods consider the data alone, and, in general, do not incorporate prior engineering knowledge. %
Some examples include multilayer perceptions \cite{MANOBEL20181015}, random forests \cite{pandit2019comparison}, and support vector machines \cite{ouyang2017modeling}. %
While these tools have proved effective in various machine learning tasks, many require stringent validation procedures, as the flexibility of the algorithms can easily lead to over-parametrised models in wind turbine applications -- as discussed in \cite{ROGERS20201124}. %

To combat the issues of overtraining, one option considers Gaussian Process (GP) regression \cite{rasmussenGP,Papatheou2015,papatheou2017performance,ROGERS20201124}. %
GPs relieve the need for validation as they are naturally self-regularising through the Bayesian Occam's razor \cite{rasmussen2001occam}; that is, training/optimisation will find the minimally-complex model given the observations in the training set. %
While GPs are typically referred to as nonparametric, a parametrised mean function (e.g.\ a sigmoid function) can be defined in the \textit{prior} of the model. In general terms, the Bayesian formulation allows for the natural inclusion of engineering knowledge of the expected functions, without the need to specify the function directly \cite{rasmussenGP}. %
As a result, GP regression can be viewed as a middle ground between purely data-based methods, and those that are based on engineering knowledge.

\subsection{Power-curtailments}\label{s:curts}

There is an established literature concerning power curtailments in wind energy; for example,~\cite{hur2014curtailment,waite2016modeling,bontekoning2017analysis,fan2015analysis,luo2016wind}. %
Generally, the literature considers physics-based simulation techniques for prediction, or control procedures to \textit{enforce} curtailment -- as opposed to data-driven models of wind-power measurements. %
For example, \citet{hur2014curtailment} present a wind farm controller to adjust the power generated by turbines while considering the requirements of the grid (for a simulated wind farm). %
It is shown that, by considering the entire wind farm in a control system, the output-power can be curtailed more effectively, such that turbines with high wind-speeds compensate for those with lower wind-speeds. %
\citet{bontekoning2017analysis} present an algorithm to determine the available power of a wind farm during curtailment, when considering the \textit{reduced wake-effect}. %
This phenomenon occurs when a turbine is curtailed, leading to a reduced wake for downstream turbines; in turn, this leads to an apparent increase in the available power. %
A physics-based model is used to adjust calculations of the available power during curtailment interactions. %
A number of papers (e.g. \citet{fan2015analysis}, \citet{luo2016wind}) have analysed the history of power-curtailments for wind energy in China, to establish potential solutions and improve the utilisation of the available resources. %
A range of technical, planning, and policy-making strategies are proposed, highlighting the importance of understanding the expected curtailments when planning wind farm projects. %

For data-driven power curve models, the curtailment data are typically considered as outliers, and removed during pre-processing; %
this is because the typical concern is to characterise \textit{ideal} operation. %
While the removal of these data makes a regression model simpler, the outlier analysis is non-trivial; %
for example, \citet{MANOBEL20181015} flag and remove outliers using a threshold based on a Gaussian Process regression, while %
\citet{marvuglia2012monitoring} de-noise the data using kernel principal component analysis. %
Alternatively, \citet{MARCIUKAITIS2017732} use the quartile/interquartile range over windowed inputs to detect and remove outliers, while %
\citet{Papatheou2015} use \textit{labels} for weekly subsets of data, provided by an expert, to remove measurements that do not correspond to ideal operation. %

\subsection{Why model power-curtailments?}\label{s:why}
While it is logical to remove curtailment data when modelling an ideal wind-power relationship, %
it is desirable to consider these `outliers' in critical applications -- namely, \textit{monitoring}, and \textit{forecasting}. %
In data-driven monitoring \cite{farrar2012structural}, the model should approximate all the variations of the permitted \textit{normal} condition to inform reliable novelty detection. %
If the model represents ideal operation only, measurements corresponding to acceptable curtailments (via control interactions) will be flagged as abnormal. %
Such a monitoring regime would lead to a large number of false positives; a recognised issue in the turbine monitoring literature~\cite{YANG2013365}. %
On the other hand, accurate curtailment modelling should prove useful within reliable forecasting frameworks. That is, if a model considers all of the expected measurements \textit{in-situ}, it should be more informative than a model of ideal-operation only; i.e. power predictions prove more conservative if curtailment data are considered. It should be noted, however, that the proposed model can only approximate curtailments that have been previously observed.

Finally, if curtailment data are modelled rather than removed, they can be naturally separated using the model itself, instead of a separate outlier analysis procedure. %
As discussed, the process of outlier removal proves far from trivial~\cite{MANOBEL20181015,marvuglia2012monitoring,MARCIUKAITIS2017732,Papatheou2015}. %

\section{Contribution}

A novel algorithm is proposed to \textit{model} curtailments in wind turbine power curves. %
The method offers an alternative to the conventional approach, which filters out the associated data. %
The algorithm expands on previous work concerning Gaussian processes (GP) \cite{ROGERS20201124,Papatheou2015,papatheou2017performance} by inferring an overlapping mixture-model of GP components -- introduced by \citet{lazaro2012overlapping}. %
An alternative (parametrised) mean function is suggested (for the GPs) that is scalable, and therefore suited to represent the expected functions for curtailed data. %
This choice of mean function allows for the inclusion of prior engineering knowledge and leads to interpretable hyperparameters. %
For each component (i.e.\ power curve) in the mixture of regression models, input-dependent (heteroscedastic) noise is approximated (according to \citet{kersting2007most}), an important consideration for probabilistic models of wind turbine power data \cite{ROGERS20201124}.

\subsection{Layout}

\Cref{s:data} introduces the SCADA dataset and the issues associated with modelling operational (curtailed) measurements. %
\Cref{s:GP_PC} summarises conventional Gaussian process regression for power curve modelling and introduces a novel parametrised mean function, as well as methods to approximate input-dependent noise for curtailment data. %
\Cref{sec:OMGP} describes the Overlapping Mixture of Gaussian Processes (OMGP) for power curve modelling, combined with ideas from \Cref{s:GP_PC}. 
\Cref{s:results} applies the model to \textit{in situ} operational SCADA data, and proposes methods for population-based monitoring with the OMGP. %
\Cref{s:conc} offers concluding remarks. %

\section{Operational Wind Farm Data: Population-based Monitoring}\label{s:data}

This work considers a SCADA dataset, recorded from an operational wind farm owned by Vattenfall, originally presented in \cite{papatheou2017performance}. For confidentiality reasons, information regarding the specific type, location, and number of turbines cannot be disclosed. The data were recorded from a farm containing the same model of turbine, over a period of 125 weeks~\cite{Papatheou2015,papatheou2017performance}. %
Observations consists of the mean \textit{power} produced and measured \textit{wind speed} over ten minute intervals. %
Sub-samples of this dataset are shown in \Cref{fig:ideal_data,fig:curtailed_data}. %

Primarily, the suggested method considers a population-based approach to performance monitoring -- associated with population-based structural health monitoring (PBSHM) \cite{PBSHMMSSP1,PBSHMMSSP2,PBSHMMSSP3}. %
That is, data from a population (the wind farm) are considered to infer a model (the power curve) that is representative of the group -- this general model is referred to as the \textit{form} in PBSHM \cite{PBSHMMSSP1}. %
To reiterate: robust and accurate models of \textit{in situ} population data are required to monitor the wind farm. 

\subsection{Dataset details}
For the SCADA data analysed in this work, the observations are \textit{unlabelled}; %
i.e.\ records of the operational, environmental, or damage condition are not available. %
Considering \Cref{fig:curtailed_data}, this fact implies that there is no ground truth to indicate which underlying function generated each sample: (i) normal operation, (ii) $\approx$ 50\% curtailment, (iii) or zero-power\footnote{While the zero-power trend is not a typical curtailment, is it considered here as a function whose data are typically filtered out before modelling. Additionally, the data are interesting to consider, as they differ functionally from other trends in the measurements.}. %
As such, when modelling the curtailments, labels to associate data with wind-power relationships (i - iii) are unobserved and must be represented as latent variables. %
It is important to note: if labels were available (in a control log, for example) they should be \textit{observed} variables in the model. %
In the absence of labelling for functions (i-iii), the model must allocate observations in an \textit{unsupervised} manner, which proves non-trivial (consider outlier analysis procedures from previous work \cite{MANOBEL20181015,marvuglia2012monitoring,MARCIUKAITIS2017732,Papatheou2015}). %

\begin{figure}[pt]
\centering\includegraphics[width=\linewidth]{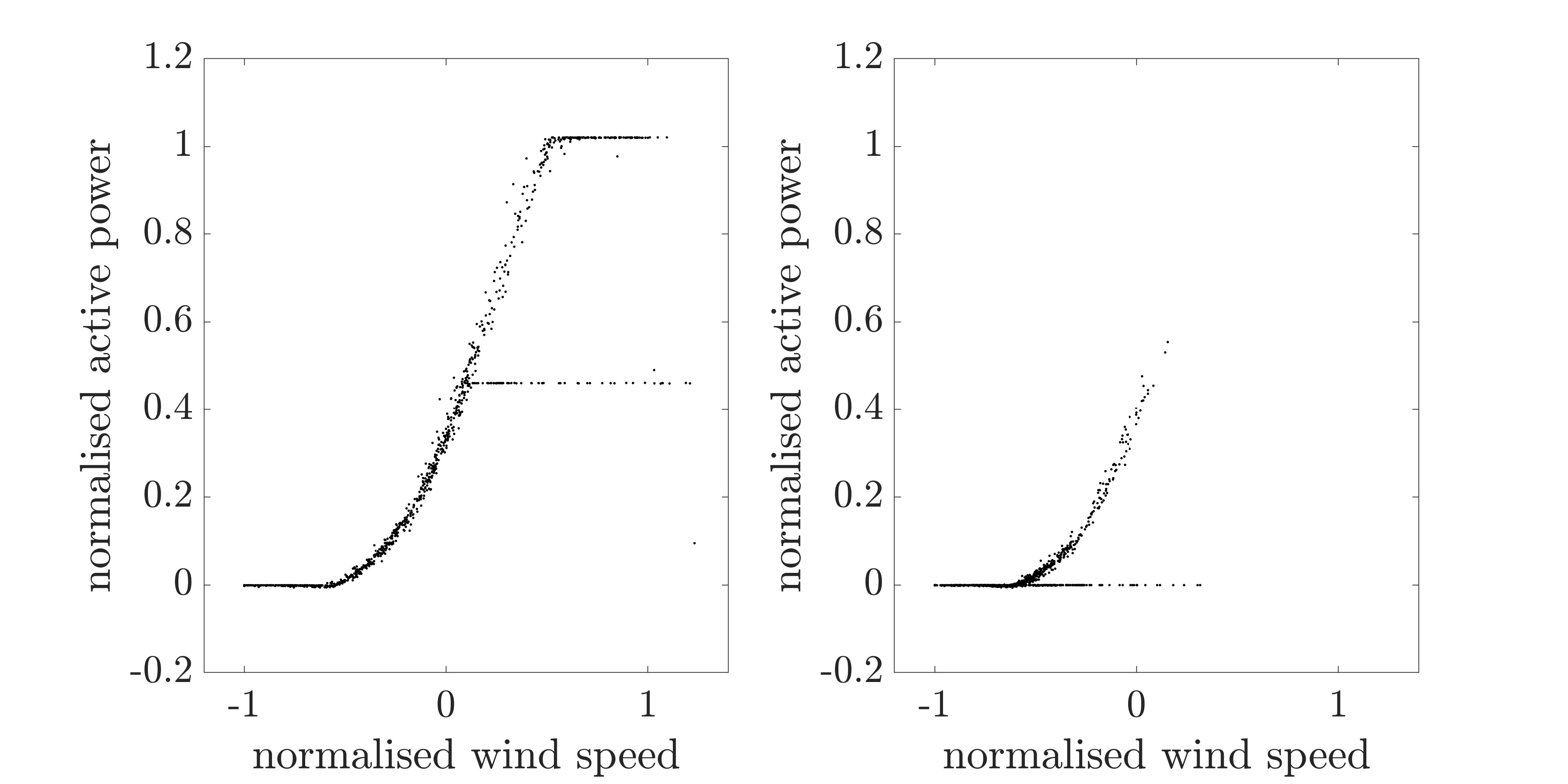}
\caption{Examples of weekly data subsets, measured from individual turbines.}\label{fig:weekly}
\end{figure}

To clarify, weekly subsets of data are presented in Figure~\ref{fig:weekly}; notice that each set can be associated with more than one operational condition~(i-iii). %
While separate trends are visually clear, manually labelling each point with the \textit{ground-truth} is infeasible. %
For example, it is clear that data represent  normal (i) and 50\% curtailment (ii) in the left of Figure~\ref{fig:weekly}; however, it becomes difficult to assign measurements to functions as they overlap. %
Likewise, while certain data clearly correspond to zero-power (iii) in the right of Figure~\ref{fig:weekly}, it is unclear if the remaining data correspond to 50\% curtailment (ii) or normal operation~(i). %

Conveniently, the labels can be modelled as a latent random variable. %
In turn, a predictive distribution can associate `soft-labels' with the data, such that a (non-zero) likelihood associates measurements with each of the underlying functions (i-iii). %

\subsection{Data selection}\label{s:selection}
While this work aims to represent more realistic measurements from an operational wind farm, it should be clarified that preprocessing steps are still required. %
The study here primarily considers data from a subset of seven turbines over a period of nine weeks (as well as four alternative turbines over seven weeks, for validation). %
Very sparse outliers are removed via a standard K-nearest-neighbour approach \cite{murphy2012machine}. %
The subsets of data were selected as they contain three trends of data (i-iii). %
It is acknowledged, however, that alternative curtailments can occur, relating to different levels of limited power. %
Some examples of alternative functions from different turbines are demonstrated in~\Cref{s:more}. %

\section{Gaussian Processes to Model Curtailed Power Curves}\label{s:GP_PC}
Before introducing the overlapping mixture model (as well as heteroscedastic updates) it is useful to summarise conventional GP regression. %
In this application, wind speed measurements correspond to the inputs $x_i$, while power measurements correspond to the outputs $y_i$. %
Given a set of $N$ training data, $\mathcal{D} = \left\{x_i,y_i\right\}_{i=1}^N = \left\{\bm{x},\bm{y}\right\}$, the predictive distribution of the power output $y_*$ for a new measurement of wind speed $x_*$ is inferred. %
Following a probabilistic approach, the power curve is modelled by some noiseless latent function $f(x_i)$, plus an independent noise term $\epsilon_i$,

\begin{align}
y_i = f(x_i) + \epsilon_i \label{eq:prob_reg}
\end{align}

\noindent Rather than inferring the parameters of a function $f$ (as with conventional parametric regression) a GP prior is placed over the functions directly. %
A Gaussian prior is also assumed for the noise term $\epsilon_i$ (the other latent variable). %
Using a Bayesian framework, a posterior distribution over the expected functions can be obtained, once training data $\mathcal{D}$ have been observed. %
The GP prior is defined by its mean $m(x_i)$ and covariance function $k(x_i,x_j)$; while the Gaussian prior is parametrised by $\sigma$,

\begin{align}
f(x_i) &\sim GP\left(m\left(x_i\right),k\left( x_i, x_j\right)\right) \\
\epsilon_i &\sim \mathcal{N}(0, \sigma^2)
\end{align}

\noindent Over a finite and arbitrary set of inputs $\bm{x} = \left\{x_1, \ldots,x_N \right\}$, the GP is a (joint) multivariate Gaussian \cite{murphy2012machine},

\begin{align}
p(\bm{f}\mid \bm{x}) = \mathcal{N}(\bm{m},\bm{K_{xx}})
\end{align}

\noindent where $\bm{m} = \{m(x_i),\ldots,m(x_N)\}$ and $
\bm{f} = \{f(x_i),\ldots,f(x_N)\}$, while $\bm{K_{xx}}$ is the covariance matrix, such that $\bm{K_{xx}}[i,j] = k(x_i,x_j) \;\;\forall i,j \in \{1,\ldots,N\}$. %
Note: square brackets are used to index matrices and vectors when subscripts become cluttered. %

Importantly, via the mean $m(x_i)$ and covariance $k(x_i,x_j)$, the GP prior can be used to encode knowledge of the expected functions given engineering judgement (before data are observed). %
The covariance function determines the correlation between outputs $y_i$ and $y_j$ -- it determines properties such as the process variance, and smoothness \cite{lazaro2012overlapping}. %
A popular (and relatively interpretable) choice of $k(\cdot)$ is the squared-exponential function (which is used here),

\begin{align}
k(x_i, x_j) = \sigma_f^2\exp{\left\{-\frac{1}{2l^2}(x_i - x_j)^2\right\}} \label{eq:sq_exp}
\end{align}

\noindent where $\sigma_f$ is the process variance, defining variance of the expected functions about the mean, and $l$ is the length scale, which determines the rate at which the correlation between outputs decays across the input space (smoothness). %

Since the GP is flexible enough to model \textit{arbitrary} trends \cite{murphy2012machine}, a zero-mean function is typically assumed \cite{rasmussenGP,lazaro2012overlapping} such that $m(x_i) = 0$; %
this is usually (somewhat) justified by subtracting the sample mean and standard deviation from the outputs $\bm{y}$. %
However, if knowledge of the \textit{expected} functions can be encoded via an explicit/parametrised mean (even approximately) this should be included \cite{PBSHMMSSP1}\footnote{It is acknowledged, however, that a poor choice of prior can lead to inferior predications.}. %
With an explicit mean, the resulting algorithm can be considered \textit{semi-parametric} \cite{murphy2012machine}, such that the GP models the residuals between the data and some parametrised function $m(x_i)$ (i.e.\ the prior mean). %

\subsection{Prior knowledge of the expected functions}

As aforementioned, sigmoid functions can be used to \textit{approximate} the expected power curve relationship: they exhibit a near-linear relationship within bounds (cut-in cut-out wind speeds) and horizontal asymptotes (re. min/max power) for high and low inputs ($x_i \rightarrow \pm \infty$). %
Sigmoids have been applied to power curves in the past -- for parametric regression, e.g.\ \cite{TASLIMIRENANI2016544,lydia2014comprehensive,MARCIUKAITIS2017732}, as well as within GPs \cite{ROGERS20201124}. %
A scaled version of the soft-clip (SC) function, presented by \citet{klimek2018neural}, is suggested as an alternative for this application, %

\begin{align}
m(x_i\;;\;\beta, \bm{\alpha}) &= \frac{\alpha_1}{\beta}\log\left\{\frac{1+e^{\beta v}}{1+e^{\beta(v-1)}}\right\} \label{eq:softclip} \\[1em]
v &\triangleq \alpha_2 x_i + \alpha_3 \nonumber\\
\bm{\alpha} &\triangleq \left\{ \alpha_1, \alpha_2, \alpha_2 \right\}
\end{align}

Relating to power curves, the hyperparameters $\{\beta, \bm{\alpha}\}$ are interpretable. %
$\alpha_1$ determines the value of the horizontal (non-zero) asymptote, which corresponds to the maximum (or limited) power. %
$\beta$ controls the \textit{rate} at which the near-linear section tends to the asymptotic values (around the cut-in/cut-out wind speed). 
Finally, $\alpha_2$ \textit{scales} and $\alpha_3$ \textit{translates} the function with respect to the $x_i$ axis. %

\begin{figure}[pt]
\centering
\includegraphics[width=\textwidth]{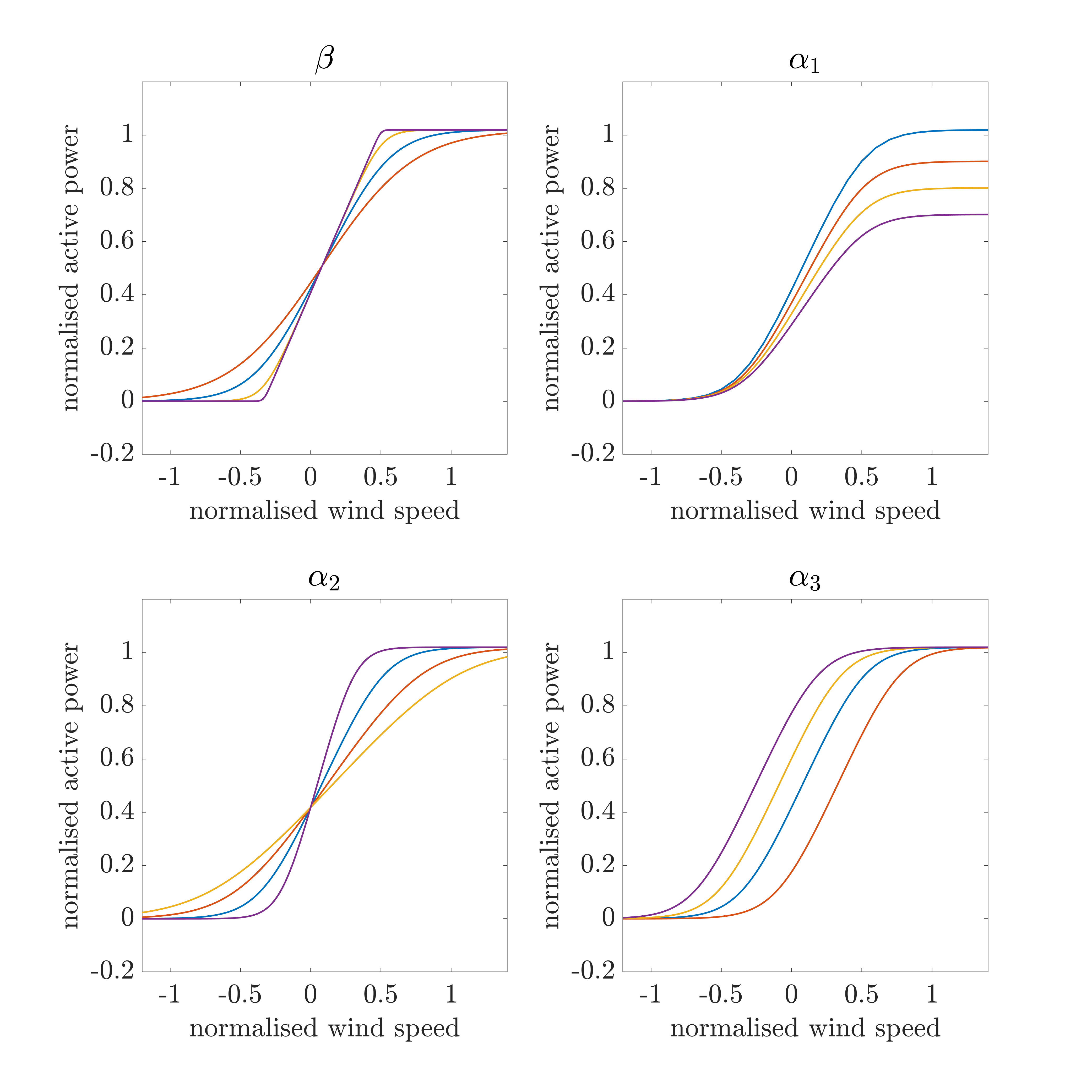}
\caption{Effects of the hyperparameters on the mean function of the prior $m(x_i; \beta, \bm{\alpha})$}\label{fig:softclip}
\end{figure}

\Cref{fig:softclip} illustrates the effects of $\{\beta, \bm{\alpha}\}$.
Importantly, control of the convergence rate via $\beta$ is particularly useful for curtailed data. %
Consider the $\approx $50\% limited trend in \Cref{fig:curtailed_data}: a sigmoid approximation would need to be scaled, such that $\alpha_1 \approx 0.5$, while $\beta$ must also increase to define \text{sharper} asymptotic behaviour. %
It is acknowledged that the zero-power trend (visible in \Cref{fig:curtailed_data}) does not resemble a soft-clip function. %
In fact, a linear regression would approximate these data -- a suitable component is introduced in \Cref{sec:OMGP}.

\subsection{Prediction and optimisation}

The collected hyperparameters of the model (associated with the mean and kernel functions) are $\bm{\theta} = \left\{\beta, \bm{\alpha}, \sigma_f, l, \sigma \right\}$. Keeping these values fixed, the joint distribution between the training data $\mathcal{D} = \left\{\bm{x},\bm{y}\right\} = \left\{x_i,y_i\right\}_{i=1}^{N}$ and some previously unseen observations $\left\{\bm{x}_*, \bm{y}_*\right\} = \left\{\bm{x}_*[i], \bm{y}_*[i] \right\}_{i=1}^{M}$ (with additive noise) is multivariate Gaussian,

\begin{align}
\begin{bmatrix}
    \bm{y}\\\ \bm{y}_*
    \end{bmatrix} &\sim \mathcal{N}\begin{pmatrix}
    \begin{bmatrix}
    \bm{m}\\ \bm{m}_*
    \end{bmatrix},
    \begin{bmatrix}
    \; \bm{K_{xx}} + \bm{R} & \bm{K_{xx_*}}\\
    \bm{K_{x_*x}} & \bm{K_{x_*x_*}} + \bm{R}_*\;
    \end{bmatrix}
    \end{pmatrix}.
    \label{eq:joint}
\end{align}
\begin{align}
    \bm{R} \triangleq \sigma^2\bm{I}_N \nonumber\\
    \bm{R}_* \triangleq \sigma^2\bm{I}_M \label{eq:noise_kernel}
\end{align}

\noindent where $\{\bm{R},\bm{R}_*\}$ define the \textit{noise kernels}, such that $\bm{I}_N$ denotes an $N \times N$ identity matrix, and $\bm{I}_M$ denotes an $M \times M$ identity matrix. Continuing similar notation, $\bm{m}_* = \left\{m\left(\bm{x}_*[i]\right)\right\}_{i=1}^{M}$ denotes the mean vector for the new observations. %

According to the standard identity for conditioning a joint Gaussian distribution \cite{murphy2012machine,rasmussenGP}, the predictive equations can be defined,

\begin{align}
p\left(\bm{y}_* \mid \bm{x}_*, \mathcal{D}\right) &= \mathcal{N}\left(\bm{\mu}_*, \bm{\Sigma}_*\right) \label{eq:predict}\\
\bm{\mu}_* &\triangleq \bm{m}_* + \bm{K_{x_*x}}\left(\bm{K_{xx}} + \bm{R}\right)^{-1} \left( \bm{y} - \bm{m}\right) \nonumber\\
\bm{\Sigma}_* &\triangleq \bm{K_{x_*x_*}} - \bm{K_{x_*x}}\left(\bm{K_{xx}} + \bm{R}\right)^{-1} \bm{K_{xx_*}} + \bm{R}_* \nonumber 
\end{align}

\noindent i.e.\ the mean of the posterior predictive function is $\mathbb{E} \left[ \bm{y}_* \right] = \bm{\mu}_*$, and the variance about that mean is $\mathbb{V}\left[\bm{y}_*\right] = \textrm{diag}(\bm{\Sigma}_*)$ (ignoring cross-terms). 

Until this point, the hyperparameters $\bm{\theta} = \left\{\beta, \bm{\alpha}, \sigma_f, l, \sigma \right\}$ have been fixed. %
In practice, these are (typically) optimised through empirical Bayes \cite{murphy2012machine}, i.e.\ a type-II maximum likelihood \cite{rasmussenGP}, see~\ref{a:typeII} for details. %

\subsection{Heteroscedastic updates: Estimating input-dependent noise}\label{s:hetGP}

\begin{figure}
\includegraphics[width=\textwidth]{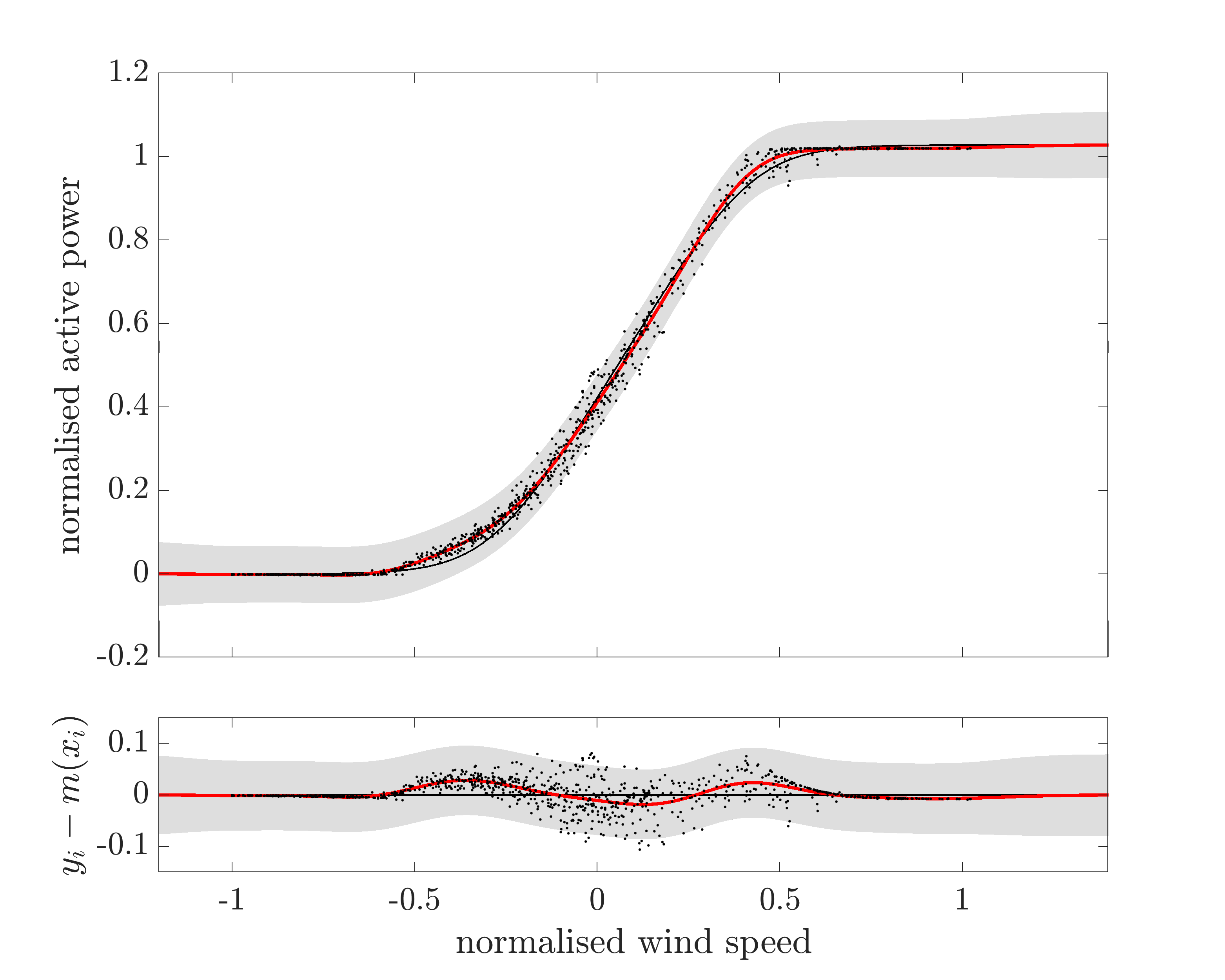}
\caption{Homoscedastic GP regression of the ideal power curve. The model in the data space (top) and the zero-mean transformed space (bottom). The black line shows the prior mean $\bm{m}$, the red line shows the predictive mean $\mathbb{E}[\bm{y}_*]$, and the shaded region shows three-sigma of the predictive variance $\mathbb{V}[\bm{y}_*]$.}\label{fig:homoGP}
\end{figure}

Currently, the noise term $\epsilon_i$ in \cref{eq:prob_reg} has been governed by a single hyperparameter  $\sigma$. When squared, $\sigma$ defines the noise variance; in turn, this defines the noise kernel $\bm{R}$ (\cref{eq:noise_kernel}). This setup enforces the assumption that the noise variance is \textit{constant} across the input domain, leading to a \textit{homoscedastatic} GP -- that is, the \textit{noise variance} does not change over $x_i$. %
To demonstrate, \Cref{fig:homoGP} depicts the homoscedastic GP learnt from the ideal data (in \Cref{fig:ideal_data}). %
The model behaves as expected: %
the mean function of the prior approximates the relationship as far as possible, while the GP models the residual between this prior and the data. To highlight this effect, the residual modelled by the GP can be visualised in the zero-mean transformed space, that is $[y_i - m(x_i)]$, \Cref{fig:homoGP}. %

While the expected function $\mathbb{E}[\bm{y}_*]$ is representative of the general trend, the noise variance is poorly approximated when $\sigma$ is constant. %
This is particularly apparent in the transformed space $\left(y_i - m\left(x_i\right)\right)$, where the noise (represented by the shaded area) is significantly overestimated at high/low wind speeds (towards the asymptotes) and underestimated in the near-linear (central) regions. %
In consequence, as proposed in \cite{ROGERS20201124}, it is necessary to model power curve data with input-dependent noise, via \textit{heteroscedastic} regression \cite{goldberg1998regression}. %
Specifically, the variance of the noise terms is now some function of the inputs $x_i$, such that,

\begin{align}
\epsilon_i &\sim \mathcal{N}(0,\sigma^2_i)\\
\sigma^2_i &= r(x_i)
\end{align}

\noindent The GP equations remain the same, other than (\ref{eq:noise_kernel}), which defined a homoscedastic noise kernel. %
For a heteroscedastic process, the diagonal of the noise kernel is now defined by $r(x_i)$, rather than a constant, such that, %

\begin{align}
    \bm{R} \triangleq \diag\left(\left\{r(x_1),\ldots,r(x_N)\right\}\right) \nonumber\\
    \bm{R}_* \triangleq \diag\left(\left\{r(x_{*1}),\ldots,r(x_{*M})\right\}\right) \label{eq:het_noise_kernel}
\end{align}
\noindent where the off-diagonal elements are zero, $\bm{R}$ is an $N \times N$ matrix, and $\bm{R}_*$ is an $M \times M$ matrix. %

Rather than specifying a functional form for the noise variance, an additional independent GP is used to infer the function $r(x_i)$. %
As $\sigma$ must be strictly positive, the GP models the log-noise levels, denoted $g(x_i)$, such that,

\begin{align}
\log(r(x_i)) = g(x_i) \sim GP(\mu_g, k_g(x_i, x_j) )
\end{align}

\noindent i.e.\ a GP prior with constant mean $\mu_g$ and a squared-exponential kernel. The kernel has the same form as \cref{eq:sq_exp}, with a distinct length scale and process variance, such that the hyperparameters of the noise-process are $\bm{\zeta} = \{\mu_g, \sigma_g, l_g \}$. %

The training points for the $g$-process can have arbitrary locations; in this case, it is convenient that they coincide with the $f$-process, such that ${\bm{g} = \{g(x_i), \ldots, g(x_N)\}}$. %
Since the noise process has been introduced as additional (conditionally-independent) latent variables $\bm{g}$, the predictive distribution for $\bm{y}_*$ (previously \cref{eq:predict}) is extended to \cite{kersting2007most},

\begin{align}
p\left(\bm{y}_* \mid \bm{x}_*, \mathcal{D}\right) = \int \int p\left(\bm{y}_* \mid \bm{x}_*, \bm{g}, \bm{g}_*, \mathcal{D}\right)p(\bm{g}, \bm{g}_*\mid \bm{x}_*, \mathcal{D}) d\bm{g}d\bm{g}_*
\end{align}

\noindent Fixing $\{\bm{g}, \bm{g}_*\}$, the predictive distribution $p\left(\bm{y}_* \mid \bm{x}_*, \bm{g}, \bm{g}_*, \mathcal{D}\right)$ is the same as before -- with \cref{eq:het_noise_kernel} defining the noise kernel $\bm{R}$. %

Unfortunately, the term $p(\bm{g}, \bm{g}_*\mid \bm{x}_*, \mathcal{D})$ is problematic, as the integral is intractable. %
Various approximations of the integral can be implemented; including Monte-Carlo approximations, as well as variational inference \cite{lazaro2011variational,ROGERS20201124,blei2017variational}. %
A simple (and computationally-inexpensive) point-wise approximation of $\bm{g}$ is utilised here. %
This approach is convenient, since input-dependent noise can be implemented as an \textit{update} step following inference of the OMGP (outlined in \Cref{sec:OMGP}). %
In this case, the approximation was found to be representative of input-dependent noise for the power curve data. %

Specifically, according to \citet{kersting2007most}, the \textit{most likely} estimate of the target noise levels is assumed for the $g$-process, such that,

\begin{align}
p\left(\bm{y}_* \mid \bm{x}_*, \mathcal{D}\right) &\approx p\left(\bm{y}_* \mid \bm{x}_*, \bm{\hat{g}}, \bm{\hat{g}}_*, \mathcal{D}\right) \\[1em]
\{\bm{\hat{g}}, \bm{\hat{g}}_*\} &\triangleq \argmax_{\{\bm{\hat{g}}, \bm{\hat{g}}_*\}}\left\{p(\bm{g}, \bm{g}_*\mid \bm{x}_*, \mathcal{D})\right\}
\end{align}

\noindent i.e.\ \textit{most} (all) of the density of $p(\bm{g}, \bm{g}_*\mid \bm{x}_*, \mathcal{D})$ is assumed to be concentrated around the mode $\{\bm{\hat{g}}, \bm{\hat{g}}_*\}$ \cite{kersting2007most}.

\subsubsection{Optimisation of the noise process}

To obtain point-wise estimates of $\bm{g}$, a homoscedastatic process is initially learnt by type-II ML -- denoted $G_1$ -- with hyperparamters $\bm{\theta}$. %
($\bm{R}$ is a constant noise kernel, as in \cref{eq:noise_kernel}.) %
Given $G_1$, an empirical estimate of the most likely noise variance can be calculated for each training observation $\left\{x_i,y_i\right\} \in \mathcal{D}$, by %
considering a sample $\tilde{y}^{(j)}_i$ from the predictive distribution of $G_1$. %
If $y_i$ and $\tilde{y}_i^{(j)}$ are viewed as two independent observations from the same underlying distribution, their arithmetic mean $0.5\left(y_i - \tilde{y}_i^{(j)}\right)^2$ is shown to be a valid approximation of the noise variance at $x_i$ \cite{kersting2007most}. %
This can be improved by taking an expectation w.r.t.\ the predictive distribution, such that \cite{kersting2007most},

\begin{align}
\log\left\{\mathbb{V}\left[y_i, G_1(x_i, \mathcal{D})\right] \right\} &\approx g^\prime_i\\
&= \log\left\{\frac{1}{s}\sum_{j=1}^{s}{0.5\left(y_i - \tilde{y}_i^{(j)}\right)^2}\right\} \label{eq:g_est}
\end{align}

\noindent here, $s$ is the sample size from the predictive distribution of $G_1$. %
A suitably large value of $s$ should lead to reasonable estimates: \citet{kersting2007most} recommend $s \geq 100$, thus, in this case, $s=100$. %
Having calculated ${\bm{g}^\prime = \{g^\prime_1,\ldots,g^\prime_N\}}$, the noise process can be learnt  -- denoted $G_2$ -- by type-II ML (given $\left\{\bm{g}^\prime, \bm{x}\right\}$) with distinct hyperparameters $\bm{\zeta}$. Then, conditioning a joint multivariate Gaussian (as before) the distribution $p(\bm{g}_* \mid \bm{x}_*, \bm{x}, \bm{g}^\prime)$ can be used to predict the (logarithmic) noise variance across the input space; in turn, defining $r(x_i)$. %

The \textit{heteroscedastic} GP -- denoted $G_3$ -- combines $G_1$ and $G_2$; %
i.e. $G_2$ models the input-dependent noise kernel according to \cref{eq:het_noise_kernel} for the $G_1$ process. %
At this point, $G_1$ is set to $G_3$ ($G_1 \leftarrow G_3$) and each step is repeated until convergence in the marginal likelihood (of the heteroscedastic process $G_3$). %
The optimisation procedure is summarised in~\ref{a:optim-HetGP}. %
Learning $\bm{g}$ in this way effectively minimises the \textit{average} distance between the target output $y_i$ and the predictive distribution of the (heteroscedastic) process $G_3$ at the training inputs~\cite{kersting2007most}. %

\subsection{Heteroscedastic regression of the ideal power curve}
\begin{figure}[pt]
\includegraphics[width=\textwidth]{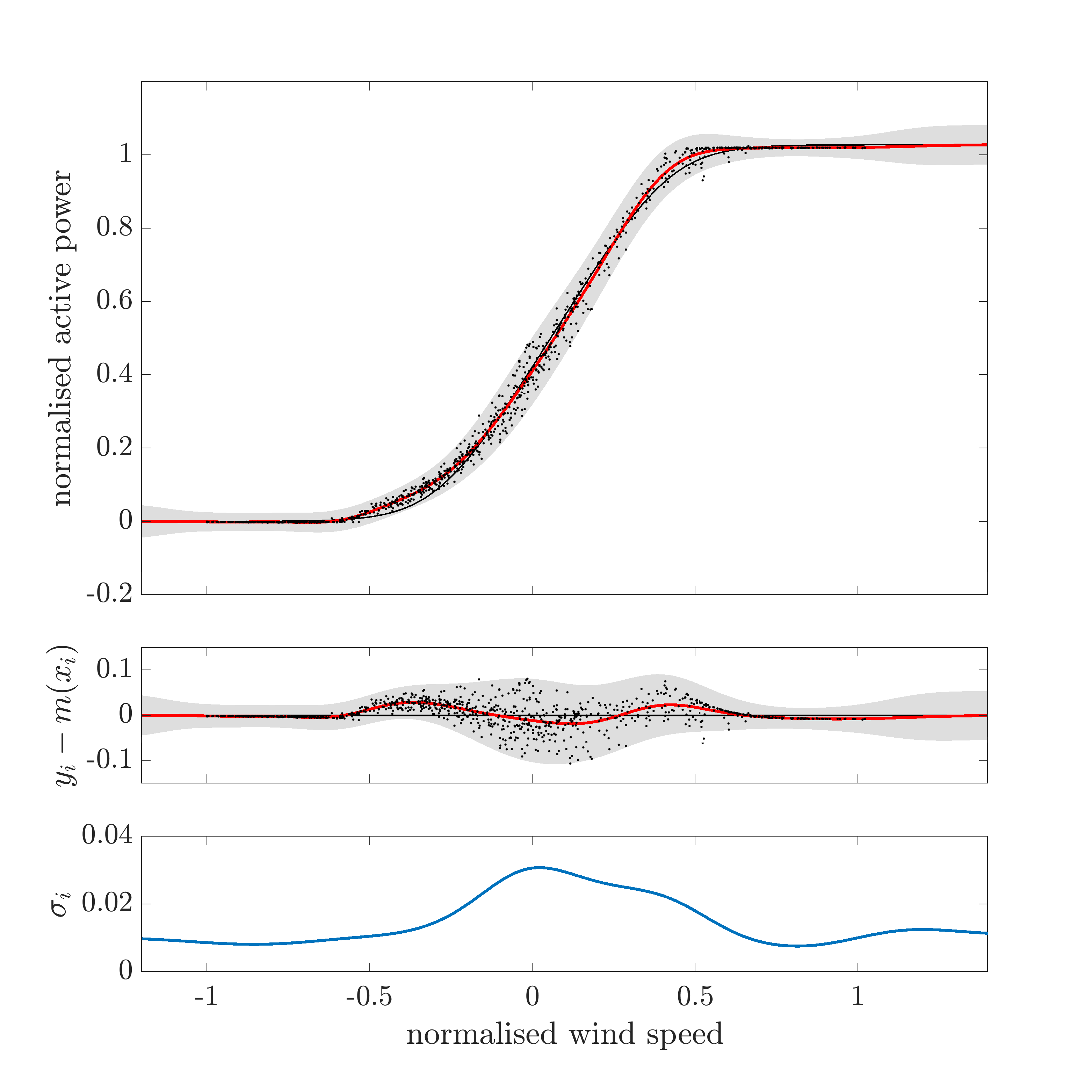}
\caption{Heteroscedastic GP regression of the ideal power curve. The model in the original space (top), the zero-mean transformed space (middle), and the expectation of the noise function $\sigma_i = \mathbb{E}\left[\sqrt{r(x_i)}\right]$ (bottom). The black line shows the prior mean $\bm{m}$, the red line shows the predictive mean $\mathbb{E}[\bm{y}_*] = \bm{\mu}_*$, and the shaded region shows three-sigma of the predictive variance $\mathbb{V}[\bm{y}_*] = \diag(\bm{\Sigma}_*)$.}\label{fig:hetGP}
\end{figure}

The optimised heteroscedastic process for the ideal data is shown in \Cref{fig:hetGP}. %
Unlike the homoscedastic example (\Cref{fig:homoGP}) the model is representative of input-dependent noise; %
to highlight this, the lower sub-plots illustrate the changing variance (shaded regions) and associated noise-levels over the inputs (blue line). %
As expected, a lower variance is associated with the tails of the sigmoid and a larger variance at the centre. %
To quantify improvements, the joint-log-likelihood of the training and test data under the model can be monitored -- this increases from $ 2.31\times 10^3$ to $3.31\times 10^3$, highlighting that input-dependant noise better approximates the variance in the data. %

The results so far, however, have shown a regression of the ideal observations only, similar to \cite{ROGERS20201124}. %
The OMGP is now introduced to model curtailed data, such as those in \Cref{fig:curtailed_data,fig:weekly}.

\section{An Overlapping Mixture of Gaussian Processes}\label{sec:OMGP}
The overlapping mixture of Gaussian processes (OMGP) model \cite{lazaro2012overlapping,tay2007smooth} is introduced to infer regression functions given curtailed power curve data. %
Here, the notation follows that of~\citet{lazaro2012overlapping}. %
Rather than a single GP, the OMGP assumes multiple latent functions to describe the data, such that,

\begin{align}
y_i^{(k)} = \left\{f^{(k)}(x_i)+ \epsilon_i\right\}_{k=1}^K
\end{align}

\noindent i.e.\ each observation is found by evaluating one of $K$ latent functions, with additive noise: %
for now, each process is homoscedastic.
As discussed, labels to assign observations to functions are unknown. %
In consequence, a latent variable is introduced to the model, $\bm{Z}$ -- this is a binary indicator matrix, such that $\bm{Z}[i,k] \neq 0$ indicates that observation $i$ was generated by function $k$. %
There is only one non-zero entry per row in $\bm{Z}$ (each observation is found by evaluating one function only). %

The likelihood of the OMGP is, therefore \cite{lazaro2012overlapping},

\begin{align}
p\left(\bm{y}\mid\left\{\bm{f}^{(k)}\right\}_{k=1}^{K},\bm{Z},\bm{x}\right) =  \prod^{N,K}_{i,k=1} p\left(y_i \mid f^{(k)}(x_i)\right)^{\bm{Z}[i,k]}
\end{align}

\noindent As with the conventional GP, prior distributions are placed over the latent functions and variables,

\begin{align}
P(\bm{Z}) &= \prod_{i,k=1}^{N,K} \bm{\Pi}[i,k]^{\bm{Z}[i,k]} \label{eq:Zprior}\\
f^{(k)}(x_i) &\sim GP\left(m^{(k)}\left(x_i\right),k^{(k)}\left( x_i, x_j\right)\right) \label{eq:fk_prior} \\
\epsilon^{(k)}_i &\sim \mathcal{N}(0, \sigma^2)
\end{align}

\noindent where \cref{eq:Zprior} is the prior over the indicator matrix, such that $\bm{\Pi}[i,:]$ is a histogram over the $K$ components for the $i^{th}$ observation, and ${\sum_{k=1}^{K}{\bm{\Pi}[i,k]} = 1}$. %
(Note, colon notation is used to index all columns or rows in a matrix.) %
The terms in \cref{eq:fk_prior} are independent GP priors over each latent function $f^{(k)}$ with distinct mean/kernel functions $\left(m^{(k)}\left(x_i\right),k^{(k)}\left( x_i, x_j\right)\right)$. To reduce the number of latent variables, the prior over the noise variances is defined by a \textit{shared} hyperparameter $\sigma$ (this is modified later, in the heteroscedastic updates). %

The collected hyperparamters for the model are $\left\{\left\{\bm{\theta}_k\right\}_{k=1}^K,\bm{\Pi}\right\}$. %
The notation $\bm{\theta}_k$ denotes a distinct set of mean/kernel function hyperparameters for the $k^{th}$ component (including the noise kernel). %
Referring back to the curtailed data in Figure~\ref{fig:curtailed_data}, it is now possible to encode prior engineering knowledge of the expected functions through the covariance, mean, and hyperparameters. %
Here, it is argued that the following are known, given prior knowledge of wind turbine power curves:

\begin{itemize} 
\item given the training data (and possibly prior knowledge of the operational conditions) it should be clear that three latent functions will be representative of the data, such that\footnote{While it is assumed $K=3$, the point-wise classification of each datum remains unknown.} $K=3$;
\item for the zero-power relationship, a linear regression (with a constant kernel) should be representative;
\item for the remaining functions (ideal and curtailed data) the soft-clip \cref{eq:softclip} appropriately describes the expected relationships.
\end{itemize}

\noindent In this setting, while $K=3$, the prior includes two independent GPs with a soft-clip mean (\cref{eq:softclip}) and squared-exponential kernel (\cref{eq:sq_exp}) function. %
These priors correspond to the ideal and curtailed curves. %
For the final component, a constant kernel is selected $k^{(3)}(x_i,x_j) = c$; this reduces the latent function to a (zero-gradient) linear regression, to approximate the zero-power data. %
To summarise, the hyperparameters (of the prior) of the model are: $\bm{\theta}_{k} = \{\beta_k, \bm{\alpha}_k, \sigma_f^{(k)}, l_k, \sigma\}_{k=1}^2$ and $\bm{\theta}_{3} = \{c,\sigma\}$. %

\subsection{A note on model assumptions}

It is important to clarify the modelling assumptions. %
While the OGMP infers labels $\mathbf{Z}$, to associate measurements to functions in an unsupervised manner, the number of functional components ($K$) and their priors must be defined in advance. %
(This concept is somewhat analogous to unsupervised learning with Gaussian Mixture models \cite{murphy2012machine}.) %
As such, while an engineer is not required to label the data, they are required to predefine an appropriate number of functions. %
Here, it is assumed that this can be determined by inspecting the static training-set (i.e.\ \Cref{fig:curtailed_data}) in an offline sense. %
In certain scenarios, however, predefining $K$ and the prior distributions is less trivial; incremental/online learning, for example. %
There are several options in this setting. %
One can select an appropriate number of components via cross-validation, considering quantities such as the Bayesian Information Criterion (BIC) or Bayes factors~\cite{murphy2012machine} -- an example of cross-validation is provided in \Cref{s:more} and \ref{a:cross-val}. %
Alternatively, $K$ could be considered as an additional latent variable, and its estimation could be included in the inference. %
Unsurprisingly inferring $K$ in this way is more involved, as presented by \citet{ross2013nonparametric}. %

It is reiterated that the training-data consider a subset of possible curtailments (described in \Cref{s:selection}). %
This consideration should not be an issue in practice, as the model is flexible in the power curves it can represent. %
To demonstrate, the OMGP is learnt for another set of curtailments, measured from four \textit{alternative} turbines in the wind farm -- the results are presented in \Cref{s:results}. %
It should be acknowledged, of course, that inference will slow down as more data (or components) are included -- a typical caveat when learning from data. %
In the context of Gaussian processes, there are a number of options; for example, sparse approximations could be explored~\cite{snelson2005sparse}. %

\subsection{Variational approximation}

Due to the latent variables $\left\{\bm{f}^{(k)}\right\}$ and $\bm{Z}$, computation of the posterior $p\left(\left\{\bm{f}^{(k)}\right\}_{k=1}^K, \bm{Z} \mid \bm{x}, \bm{y}\right)$ is now intractable; %
thus, variational inference (VI) \cite{blei2017variational} is implemented. %
Specifically, VI involves specifying an approximate density family $q(\bm{a}) \in \mathcal{Q}$ over the target conditional $p(\bm{a}\mid\bm{b})$.
The best candidate $\hat{q}(\bm{a})$ can be viewed as $q(\bm{a}) \in \mathcal{Q}$ that is \textit{closest} to the (unknown) target $p(\bm{a}\mid\bm{b})$ in terms of the KL-divergence,

\begin{align}
\hat{q}(\bm{a}) = \argmin_{q(\bm{a}) \in \mathcal{Q}} KL\left(q(\bm{a})\mid\mid p(\bm{a}\mid\bm{b})\right) \label{eq:VIKL}
\end{align}

\noindent Once found, $\hat{q}(\bm{a})$ is the best approximation of $p(\bm{a}\mid\bm{b})$ within the family $\mathcal{Q}$~\cite{blei2017variational}. %
(In this case, $\bm{a} \triangleq \left\{\left\{\bm{f}^{(k)}\right\}, \bm{Z}\right\}$ and $\bm{b} \triangleq \{\bm{y}\}$.) %
The KL divergence for \cref{eq:VIKL} is defined,

\begin{align}
 KL\left(q(\bm{a})\mid\mid p(\bm{a}\mid\bm{b})\right) &= \mathbb{E}_{q(\bm{a})}[\log{q(\bm{a})}] - \mathbb{E}_{q(\bm{a})}[\log{p(\bm{a}\mid\bm{b})}] \label{eq:KL1}\\
 &= \mathbb{E}_{q(\bm{a})}[\log{q(\bm{a})}] - \mathbb{E}_{q(\bm{a})}[\log{p(\bm{a},\bm{b})}] + \log{p(\bm{b})} \label{eq:KL2}
\end{align}{}

\noindent (Steps (\ref{eq:KL1}) to (\ref{eq:KL2}) uses log rules while expanding the conditional.) %
\cref{eq:KL2} reveals the dependence on $p(\bm{b})$, which is intractable, and why VI is needed in the first place \cite{blei2017variational}. Therefore, rather than the KL divergence (\ref{eq:KL2}), an alternative object is optimised that is equivalent to the (negative) KL divergence up to the term $\log{p(\bm{b})}$, which is a constant with respect to $q(\bm{z})$; that is,

\begin{align}
\mathcal{L}_{b}(\bm{a})&= \mathbb{E}_{q(\bm{a})}[\log{p(\bm{a},\bm{b})}] -\mathbb{E}_{q(\bm{a})}[\log{q(\bm{a})}] \label{eq:elbo1}\\
&= \int q(\bm{a}) \log{\frac{p(\bm{a},\bm{b})}{q(\bm{a})}} \d\bm{a} \label{eq:elbo}
\end{align}

\noindent This quantity is the referred to as the \textit{evidence lower bound} (elbo). From (\ref{eq:KL2}), it can be seen that \textit{maximising} this object will \textit{minimise} the KL divergence between $q(\bm{a})$ and $p(\bm{a}\mid\bm{b})$. %

Conveniently, \cref{eq:elbo} can be used to construct a lower bound on the marginal likelihood $p(\bm{b})$: %
i.e.\ rearranging \cref{eq:KL2} and substituting in (\ref{eq:elbo1}) leads to,

\begin{align}
\log{p(\bm{b})} &=  KL\left(q\left(\bm{a}\right) \;\Vert\; p\left(\bm{a}\mid\bm{b}\right)\right) + \mathcal{L}_{b} \label{eq:KL_LB}
\end{align}

\noindent Since $KL(\cdot)\geq0$ \cite{mackay2003information}, it follows that the evidence is lower-bounded by the elbo, in other words \ $\log{p(\bm{b})}\geq \mathcal{L}_{b}$. %
This inequality is useful, as $\mathcal{L}_{b}$ can be used to monitor the marginal likelihood during inference/optimisation (as with the conventional GP, \cref{eq:lml}). %
Substituting notation $\bm{a} \triangleq \left\{\left\{\bm{f}^{(k)}\right\}, \bm{Z}\right\}$ and $\bm{b} \triangleq \{\bm{y}\}$ in (\ref{eq:elbo}), leads to (\ref{eq:L_b}),

\begin{small}
\begin{align}
&\log{p(\bm{y}\mid\bm{x})} = \log \int \int p\left(\left\{\bm{f}^{(k)}\right\}, \bm{Z}, \bm{y}, \bm{x}\right) p\left(\left\{\bm{f}^{(k)}\right\}\right) p\left(\bm{Z}\right) \d\left\{\bm{f}^{(k)}\right\} \d\bm{Z} \label{eq:tru_lml}\\
&\geq \mathcal{L}_{b} = \int \int q\left(\left\{\bm{f}^{(k)}\right\}, \bm{Z}\right) \log{\frac{p\left(\left\{\bm{f}^{(k)}\right\}, \bm{Z}, \bm{y}, \bm{x}\right)}{q\left(\left\{\bm{f}^{(k)}\right\}, \bm{Z}\right) }} \d\left\{\bm{f}^{(k)}\right\} \d\bm{Z} \nonumber\\
&= \int\int q\left(\left\{\bm{f}^{(k)}\right\}, \bm{Z}\right)
\log{\frac{p\left(\bm{y}\mid\left\{\bm{f}^{(k)}\right\},\bm{Z},\bm{x}\right)p(\bm{Z}) \prod_{k=1}^K p\left(\bm{f}^{(k)}\mid \bm{x}\right)}{q\left(\left\{\bm{f}^{(k)}\right\}, \bm{Z}\right)}}\;\d\left\{\bm{f}^{(k)}\right\}\d\bm{Z} \label{eq:L_b}
\end{align}
\end{small}

A family of variational distributions $q \in \mathcal{Q}$ is now chosen to approximate $p\left(\left\{\bm{f}^{(k)}\right\}, \bm{Z} \mid \bm{x}, \bm{y}\right)$ such that a mean field assumption is implemented: i.e.\ $q$ factorises, $q\left(\left\{\bm{f}^{(k)}\right\}, \bm{Z}\right) = q\left(\left\{\bm{f}^{(k)}\right\}\right) q\left(\bm{Z}\right)$. %
In consequence, due to conjugacy, it is possible to analytically update each latent variable in turn (while keeping the others fixed) such that the bound $\mathcal{L}_b$ is maximised (with respect to that variable). %
Updates for each factor are iterated until convergence in the lower bound $\mathcal{L}_b$.\footnote{At this stage in the inference, the hyperparameters of the model $\left\{\left\{\bm{\theta}_k\right\}_{k=1}^K,\bm{\Pi}\right\}$ are fixed -- they will be optimised later.}

\subsubsection{Mean-field updates}
Firstly, assuming $q\left(\left\{\bm{f}^{(k)}\right\}\right)$ is known -- and therefore the marginals for each component $q\left(\bm{f}^{(k)}\right)= \mathcal{N}\left(\bm{\mu}^{(k)}, \bm{\Sigma}^{(k)}\right)$ -- it is possible to analytically maximise $\mathcal{L}_{b}$ with respect to $q(\bm{Z})$, by setting the derivative of the bound to zero, and constraining $q$ to be a probability density~\cite{lazaro2012overlapping},

\begin{align}
  q(\bm{Z}) &= \prod_{i=1,k=1}^{N,K} \bm{\hat{\Pi}}[i,k]^{\bm{Z}[i,k]}, \hspace{1em} \textrm{s.t.} \hspace{1em}
  \bm{\hat{\Pi}}[i,k] \propto \bm{\Pi}[i,k] \exp\left(a_{ik}\right) \label{eq:VIup1}\\
  a_{ik} &\triangleq - \frac{1}{2\sigma^2} \left(\left( y_i - \bm{\mu}^{(k)}_i \right)^2 + \bm{\Sigma}^{(k)}[i,i] \right) - \frac{1}{2} \log\left(2\pi\sigma^2\right)  \nonumber
\end{align}

\noindent where \cref{eq:VIup1} implies the approximated distribution $q(\bm{Z})$ is factorised for each sample~\cite{lazaro2012overlapping}. %

Conversely, assuming $q(\bm{Z})$ is known, $\mathcal{L}_{b}$ can maximised with respect to each latent function $q\left(\left\{\bm{f}^{(k)}\right\}\right)$,

\begin{align}
  q\left(\bm{f}^{(k)}\right) &= \mathcal{N}\left(\bm{f}^{(k)}\mid\bm{\mu}^{(k)}, \bm{\Sigma}^{(k)}\right) \label{eq:VIup2} \\[1ex]
  \bm{\Sigma}^{(k)} &\triangleq \left(\bm{K}^{-1{(k)}}_{\bm{x}\bm{x}} + \bm{B}^{(k)} \right)^{-1} \nonumber\\
  \bm{\mu}^{(k)} &\triangleq \bm{m}^{(k)}+ \bm{\Sigma}^{(k)} \bm{B}^{(k)}\left(\bm{y} - \bm{m}^{(k)}\right) \nonumber
\end{align}

\noindent where $\bm{B}^{(k)}$ is a $N\times N$ diagonal matrix (off-diagonals are zero) with elements,

\begin{align}
  \bm{B}^{(k)} = \diag\left(\left\{\frac{\bm{\hat{\Pi}}[1,k]}{\sigma^{2}}\,,\; \ldots\;,\, \frac{[\bm{\hat{\Pi}}[N,k]}{\sigma^{2}} \right\}\right) \label{eq:omgpB}
\end{align}

\noindent To find a candidate $\hat{q}$ that is closest to the true posterior, $q(\bm{Z})$ and $q\left(\left\{\bm{f}^{(k)}\right\}\right)$ are initialised from their priors, and they are iteratively updated by alternating \cref{eq:VIup1,eq:VIup2}. %
Both updates are optimal with respect to the distribution that they compute; therefore, they are guaranteed to increase the (lower bound) on the log-marginal-likelihood~\cite{lazaro2012overlapping}, \cref{eq:L_b}.

\subsubsection{Monitoring convergence: An improved lower bound}
As in \cite{lazaro2011variational}, an improved bound is used to \textit{monitor} convergence, introduced by \citet{king2006fast}. %
This object, denoted $\mathcal{L}_{bc}$, also lower-bounds the marginal likelihood, while being an upper-bound on the standard variational bound $\mathcal{L}_b$ (\cref{eq:L_b}). %
(That is, if $\mathcal{L}_b$ is subtracted from the improved bound, the result is a KL divergence -- %
as $KL(\cdot) \geq 0$, this implies that $\mathcal{L}_{bc}$ upper-bounds $\mathcal{L}_b$.) %
The bound can be defined when the term $\log \int p\left(\left\{\bm{f}^{(k)}\right\}, \bm{Z}, \bm{y}, \bm{x}\right) p\left(\bm{Z}\right) \d\bm{Z}$ -- in the true marginal likelihood, \cref{eq:tru_lml} -- is replaced with $ \int q(\bm{Z}) \log \frac{p\left(\left\{\bm{f}^{(k)}\right\}, \bm{Z}, \bm{y}, \bm{x}\right) p\left(\bm{Z}\right)}{q(\bm{Z})} \d\bm{Z}$. %
Following substitution, it is possible to integrate out $p\left(\left\{\bm{f}^{(k)}\right\}\right)$ analytically.
Alternatively, \citet{lazaro2011variational} show that it is possible to obtain the corrected bound by optimally removing $p\left(\left\{\bm{f}^{(k)}\right\}\right)$ from the standard bound.
The (implementation friendly) expression for $\mathcal{L}_{bc}$ is as follows \cite{lazaro2011variational},

\begin{align}
  &\log{p(\bm{y}\mid\bm{x})} \geq \mathcal{L}_{bc} \nonumber\\
  & = \sum_{k=1}^{K}\left(-\frac{1}{2} \big\lVert \bm{R}^{(k)\top} \backslash \left(\bm{B}^{(k)\frac{1}{2}} \left( \bm{y} - \bm{m}^{(k)} \right) \right) \big\rVert^2 - \sum_{i=1}^{N} \log\bm{R}^{(k)}[i,i] \right) \ldots \nonumber\\[1ex]
  & \quad\qquad\qquad - \textrm{KL}\left(q\left(\bm{Z}\right) \,\big\Vert\, p\left(\bm{Z}\right) \right) - \frac{1}{2} \sum^{N,K}_{i=1,k=1} \bm{\hat{\Pi}}[i,k] \log\left(2\pi\sigma^{^2}\right) \label{eq:Lvb}\\[2em]
  &\bm{R}^{(k)} \triangleq \textrm{chol}\left(\bm{I} + \bm{B}^{(k)\frac{1}{2}} \,\bm{K}^{{(k)}}_{\bm{x}\bm{x}}\, \bm{B}^{(k)\frac{1}{2}}\right) \nonumber \\[1ex]
  &\textrm{KL}\left(q\left(\bm{Z}\right) \,\big\Vert\, p\left(\bm{Z}\right) \right) \triangleq \sum_{i=1,k=1}^{K,N} \bm{\hat{\Pi}}[i,k] \log\frac{\bm{{\Pi}}[i,k]}{\bm{\hat{\Pi}}[i,k]} \nonumber
\end{align}

\noindent where $\textrm{chol}(\cdot)$ is the Cholesky decomposition and the backslash operator $\bm{A}\backslash \bm{B}$ solves the systems of linear equations $\bm{A}\bm{c}=\bm{B}$ for $\bm{c}$. The improved, tighter bound is independent of $p\left(\left\{\bm{f}^{(k)}\right\}\right)$ -- hence it can be referred to as the \textit{marginalised variational bound} \cite{lazaro2011variational}. %
In words, this implies that $\mathcal{L}_b$ is the same as $\mathcal{L}_{bc}$ -- for a given $q(\bm{Z})$ -- when an optimal choice for $p\left(\left\{\bm{f}^{(k)}\right\}\right)$ is made \cite{lazaro2012overlapping}. %
In consequence, the bound is more stable over different hyperparameter values \cite{king2006fast}, and it is more efficient when optimising $\left\{\left\{\bm{\theta}_k\right\}_{k=1}^K,\bm{\Pi}\right\}$ through type-II ML (the optimisation scheme is outlined below).

\subsubsection{Optimisation of hyperparameters}
A variational inference and Expectation Maximisation (EM) scheme is implemented \cite{blei2017variational}. %
The strategy iteratively updates the \textit{approximate} (factorised) posterior and then optimises the hyperparameters of the model, while the (improved) lower bound $\mathcal{L}_{bc}$ on the marginal likelihood is maximised. The EM steps are repeated until convergence:

\begin{enumerate}
\item E-step: mean field updates -- iterate \cref{eq:VIup1,eq:VIup2} until convergence in $\mathcal{L}_{bc}$ (or $\mathcal{L}_{b}$), hyperparameters are fixed.
\item M-step: optimise the lower bound $\mathcal{L}_{bc}$ w.r.t.\ all hyperparameters until convergence,
$$\left\{\hat{\left\{\bm{\theta}_k\right\}}_{k=1}^K,\hat{\bm{\Pi}}\right\}= \underset{\left\{\left\{\bm{\theta}_k\right\}_{k=1}^K,\bm{\Pi}\right\}}{\argmax}\bigg\{\mathcal{L}_{bc}\bigg\}$$
the distribution $q(\bm{Z})$ is kept fixed. %
\end{enumerate}

\noindent Having initialised each component from the prior, steps 1 and 2 are iterated until convergence in $\mathcal{L}_{bc}$ (of the M-step).

\subsubsection{Predictive equations}

Having learnt the OMGP, it can be used to estimate the latent variables and functions. %
These predications are critical in the context of performance monitoring: i.e.\ for a given measurement of wind speed $x_i$, the OMGP can predict the power output $y_i$, and classify the trend (or curtailment) $k \in \{1,\ldots,K\}$. 
The posterior predictive likelihood given the unseen inputs $\bm{x}_*$ is,
\begin{align}
  p(\bm{y}_*\mid\bm{x}_*,\mathcal{D}) &\approx \sum^K_{k=1} \bm{\Pi}[*,k] \int p\left(\bm{y}_*\mid \bm{f}^{(k)},\bm{x}_*, \mathcal{D}\right)q\left(\bm{f}^{(k)} \mid \mathcal{D}\right) \d \bm{f}^{(k)}\\
  &=\sum_{k=1}^K \bm{{\Pi}}[*,k] \; \mathcal{N}\left(\bm{y}_*\mid\bm{\mu}^{(k)}_*,\bm{\Sigma}_{*}^{(k)}\right) \label{eq:PPL}\\[1em]
  \bm{\mu}^{(k)}_* &\triangleq \bm{m}_*^{(k)} + \bm{K}_{\bm{x}_*\bm{x}}^{(k)}\left(\bm{K}_{\bm{x}\bm{x}}^{(k)} + \bm{B}^{(k)-1} \right)^{-1} \left(\bm{y} - \bm{m}^{(k)}\right) \nonumber\\
  \bm{\Sigma}_{*}^{(k)} &\triangleq \bm{K}_{\bm{x}_*\bm{x}_*}^{(k)} - \bm{K}_{\bm{x}_*\bm{x}}^{(k)}\left(\bm{K}_{\bm{x}\bm{x}}^{(k)} + \bm{B}^{(k)-1} \right)^{-1} \bm{K}_{\bm{x}\bm{x}_*}^{(k)} + \bm{R}_*^{(k)} \nonumber \\
  \bm{R}_*^{(k)} &\triangleq \sigma^2\,\bm{I}_M \label{eq:omgpR}
\end{align}

\noindent The prior mixing proportion for new observations $\bm{{\Pi}}[*,k]$ is a fixed hyperparameter, weighting each component equally \textit{a priori}, such that $\bm{{\Pi}}[*,k] = 1/K$. %
Interestingly, the predictive equations for the OMGP are similar to the conventional GP (\cref{eq:predict}), however, the noise component for the training data ($\bm{B}^{(k)-1}$) is scaled according to $\bm{\hat{\Pi}}[i,k]^{-1}$ \cite{lazaro2012overlapping}. 
Thus, the noise component effectively \textit{weights} the contribution of each observation in $\mathcal{D}$ to its posterior predictive component in the mixture.

Another useful prediction categorises observations according to the most likely component $k$. For the training data $\mathcal{D}$, this is simply the \textit{maximum a posteriori} (MAP) estimate, given the approximated posterior (\ref{eq:VIup1}),

\begin{equation}
  \hat{k}_i = \underset{k}{\textrm{argmax}}\left\{\bm{\hat{\Pi}}[i,k]\right\}
\end{equation}

\noindent For the test-data (i.e.\ weekly wind-power data $\{\bm{x}_*,\bm{y}_*\}$) the posterior predictive class component ${k}_*$ is,

\begin{align}
  p(k_*\mid\bm{x}_*,\bm{y}_*,\mathcal{D}) &= \frac{p(\bm{y}_*\mid\bm{x}_*,k,\mathcal{D})\,\bm{{\Pi}}[*,k]}{p(\bm{y}_*\mid\bm{x}_*, \mathcal{D})} \label{eq:bayes_rules_classify}
\end{align}

\noindent where the denominator (evidence) was defined in (\ref{eq:PPL}), and the numerator is,

\begin{align}
  p(\bm{y}_*\mid\bm{x}_*,k_*,\mathcal{D})\,p(k_*) &\triangleq \mathcal{N}\left(\bm{y}_*\mid\bm{\mu}^{(k)}_*,\bm{\Sigma}_{*}^{(k)}\right) \bm{{\Pi}}[*,k] 
\end{align}

\noindent the MAP class component $\hat{k}_*$ can then be defined,

\begin{align}
  \hat{k}_* &= \underset{k_*}{\textrm{argmax}}\left\{p(k_*\mid\bm{x}_*,\bm{y}_*,\mathcal{D})\right\} \label{eq:classify}
\end{align}

\noindent Note, classifying new data according to $\hat{k}_*$ is only possible when \textit{both} $\bm{x}_*$ and $\bm{y}_*$ have been observed. %
This implies that predictions using \cref{eq:classify} should be used in certain monitoring applications (as demonstrated in the results). %

\subsubsection{Input dependent noise approximations for the OMGP}

At this stage, it is possible to apply heteroscedastic updates to the OMGP, according to the method in \Cref{s:hetGP}. %
In this case, for each $k^{th}$ component, the noise variance is now considered a function of the inputs,

\begin{align}
\sigma_i^{(k)^2} &= r^{(k)}(x_i) \\[1em]
\log r^{(k)}(x_i) = g^{(k)}(x_i) &\sim GP(\mu_g^{(k)},k_g^{(k)}(x_i,x_j))
\end{align}

\noindent i.e.\ there are $K$ GPs (with hyperparameters $\bm{\zeta}_k = \{\mu^{(k)}_g, \sigma^{(k)}_g, l^{(k)}_g \}$) to describe input-dependent noise for each function in the mixture -- rather than a single, shared hyperparameter $\sigma$.

Again, the predictive \cref{eq:PPL} remains similar, where the noise kernels are updated. %
In this case, $\bm{B}^{(k)}$ (from \cref{eq:omgpB}) becomes,

\begin{align}
\bm{B}^{(k)} = \diag \left(\left\{\frac{\bm{\hat{\Pi}}[\mathcal{I}^{(k)}_1,k]}{r^{(k)}(x_{\mathcal{I}^{(k)}_1})}\,,\; \ldots\;,\, \frac{[\bm{\hat{\Pi}}[\mathcal{I}^{(k)}_N,k]}{r^{(k)}(x_{\mathcal{I}^{(k)}_N})} \right\}\right)
\end{align}

\noindent where the indices $\bm{\mathcal{I}}_k=\{\mathcal{I}^{(k)}_1,\ldots \mathcal{I}^{(k)}_N\}$ correspond to observations in $\mathcal{D}$ whose MAP label is $k$. %
Formally, $\{x_i, y_i\}_{i\in\bm{\mathcal{I}}_k} \in \mathcal{D}$, where $\hat{k}_{i\in \bm{\mathcal{I}}_k} = k$.
Additionally, $\bm{R}_*^{(k)}$ from \cref{eq:omgpR} is updated,

\begin{align}
\bm{R}_*^{(k)} &\triangleq \diag\left(\left\{r^{(k)}(x_{*1}),\ldots,r^{(k)}(x_{*M})\right\}\right)
\end{align}

In summary, to approximate the noise-process for each component, the training data are split into $K$ subsets, according to the MAP classification (\ref{eq:classify}) given the homoscedastic OMGP and the training data. %
That is, the noise-processes are approximated for each component, given the training data that are associated with that component (according to $\hat{k}_i$) and the framework outlined in \Cref{s:hetGP}.


\section{Results}\label{s:results}
In total, 8900 observations were sampled from the wind farm data, corresponding to a (selected) subset of seven operational turbines over nine weeks. %
As aforementioned, three trends are present in these data; additional curtailments may be observed in practical data, and can be included in the OMGP if necessary -- an alternative example is provided in \Cref{s:more}. %
The data are shown in \Cref{fig:curtailed_data,fig:weekly}. %
Approximately 1/3 ($N=2980$ observations) were using for training here, and the remaining data ($M=5920$ observations) were used as an independent test-set. %

\begin{figure}[pt]
\includegraphics[width=\textwidth]{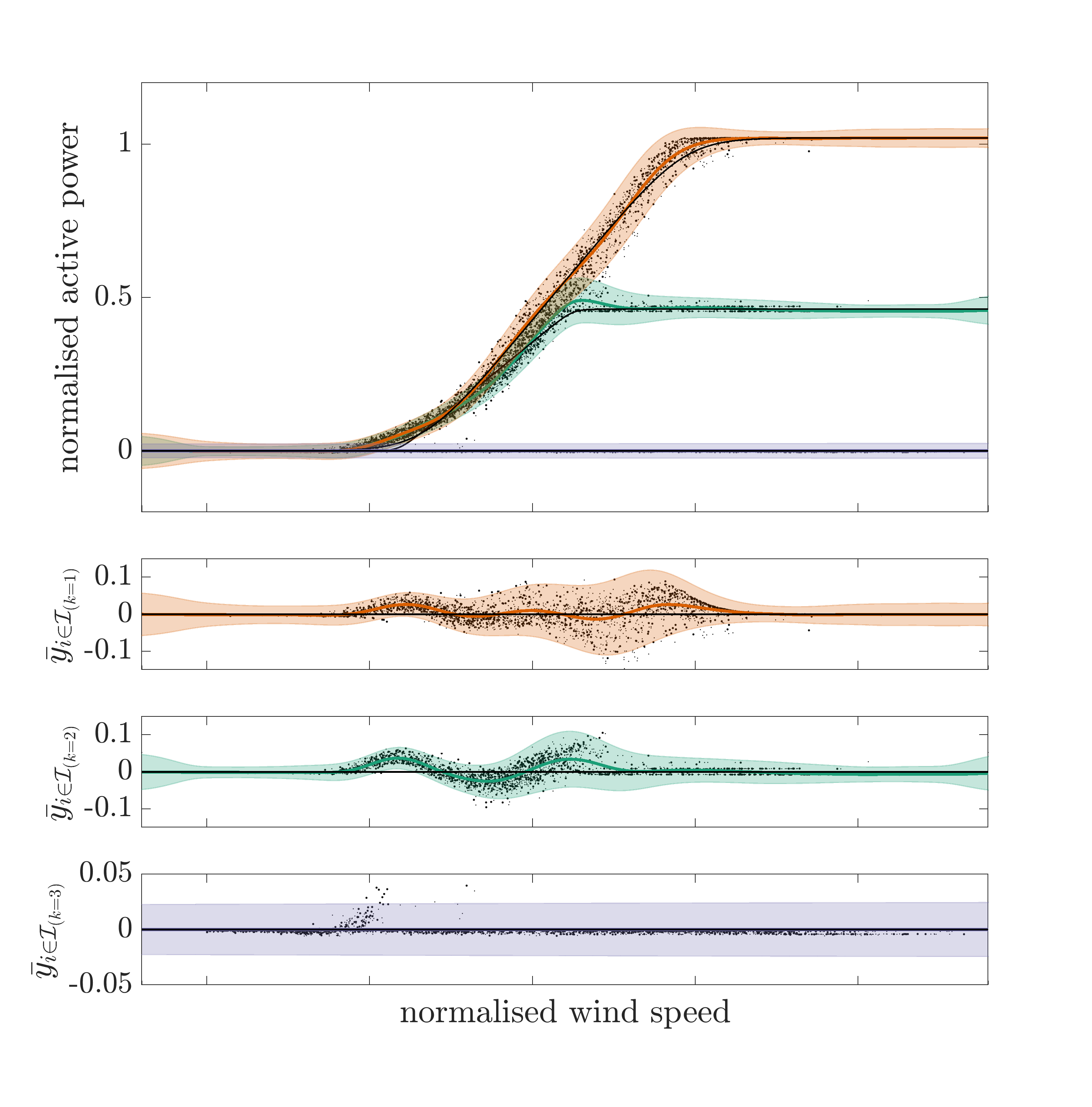}
\caption{Heteroscedastic OMGP regression of curtailed power curve data. The mixture model in the original space (top), and each component in the zero-mean transformed space, i.e.\ $\bar{y}_i = y_i - m^{(k)}(x_i)$ (bottom three plots). Black lines show the mean functions of the prior $\bm{m}^{(k)}$. The green, orange, and purple lines show the predictive mean $\bm{\mu}^{(k)}_*$, and shaded regions show three-sigma of the predictive variance $\diag(\bm{\Sigma}^{(k)}_*)$. Small {$\bm{\cdot}$} markers show the test set, and larger $\bullet$ markers show the training set. %
For each component, the data correspond to their MAP function, according to $\hat{k}_i$ and $\hat{k}_*$.}\label{fig:hetOMGP}
\end{figure}

OMGP regression of the curtailed data is shown in \Cref{fig:hetOMGP}. %
Given the training observations (larger {$\bullet$} markers), the model has inferred the multivalued behaviour in an unsupervised manner, including the ideal curve (orange), $\approx$ 50\% curtailment (green), and the zero-power behaviour (purple). %

Visually, the model is representative of the underlying functions, and it appears to generalise to the test data (smaller {$\bm{\cdot}$} markers). %
Importantly, the GP successfully models the residual between prior engineering knowledge (encoded in the parametrised mean, shown by the black lines in \Cref{fig:hetOMGP}) and the data. %
Generally, the heteroscedastic updates are representative. The noise levels are (perhaps) overestimated towards the asymptotes of the power curves (high and low wind speeds). %
Additionally, the noise for the zero-power trend (purple) is overestimated, as it captures some of the data associated with the ideal/curtailed data -- around negative $0.5$ normalised wind speed. %
Smaller length scales $l_g^{(k)}$ in the noise-processes $g^{(k)}$ might prove appropriate, as there is no guarantee that the parameter set $\left\{\left\{\bm{\theta}_k, \bm{\zeta}_k \right\}_{k=1}^K,\bm{\Pi}\right\}$ represents the \textit{global} minimum of the log-marginal-likelihood. %
However, following several initialisations, this realisation was the most representative (and repeatable). %

To quantify performance, the normalised mean squared-error (NMSE) and Mahalanobis squared-distance (MSD) are provided. %
As the OMGP is a mixture, each test observation is assessed against its most likely component $\hat{k}_*$. In other words, NMSE and (normalised) MSD are assessed for each function with respect to their most likely data -- the corresponding subsets are shown for each component in the (lower) plots of \Cref{fig:hetOMGP}.

\begin{align}
\hat{NMSE} = \frac{100}{M \sigma_{\bm{y}_*}^2}\left(\bm{\mu}^{(\hat{k}_*)}_*-\bm{y}_*\right)^{\top}\left(\bm{\mu}^{(\hat{k}_*)}_*-\bm{y}_*\right) \label{eq:nmse}
\end{align}

\noindent Similarly, the MSD is,
{}
\begin{align}
\hat{MSD} = \frac{1}{M} \sum^{M}_{i=1} \frac{\left(\bm{\mu}^{(\hat{k}_*)}_*[i] -\bm{y}_*[i]\right)^2}{\bm{\Sigma}^{(\hat{k}_*)}[i,i]} \label{eq:msd}
\end{align}

\begin{table}[]
\centering
\begin{tabular}{l|cc|cc}

 \hspace{1em} & \multicolumn{2}{c}{Conventional regression} & \multicolumn{2}{c}{Mixture of regressions} \\   
 \hspace{1em} & RVM & GP & OMGP & Het-OMGP \\
 \hline
 \hline
 NMSE & 47.13 & 46.97 & 0.26 & 0.26 \\
 MSD & 1.01 & 1.02 & 1.00 & 0.73
\end{tabular}\caption{Model performance metrics for the curtailed power curve data.}\label{t:metrics}
\end{table}

\Cref{t:metrics} quantifies significant improvements in representing the curtailed power data with a heteroscedastic OMGP. %
For reference, an alternative probabilistic regression is included, previously applied in the literature~\cite{jing2020wind}, the Relevance Vector Machine (RVM); implementation details are provided in~\ref{a:benchmarks}. %
It is reiterated, however: the focus is to show improvements of a mixture of regressions, rather than improvements between conventional regression models. %

The NMSE shows a marked advantage in representing the data with multiple latent functions. %
Nonetheless, the NMSE does not highlight advantages of heteroscedastic updates, since the metric (\ref{eq:nmse}) does not consider the predictive variance $\bm{\Sigma}^{(k)}_*$. %
Therefore, the (normalised) MSD in \Cref{t:metrics} highlights
 improvements when modelling input-dependent noise for the mixture%
\footnote{It is acknowledged that the MSD is less useful when assessing the \textit{fit} of the OMGP, as the \textit{error} is scaled by the predictive variance $\bm{\Sigma}^{(k)}_*$; %
thus, the MSD is used only to assess the predictive variance $\bm{\Sigma}^{(k)}_*$.}. %

As discussed, certain hyperparamters can be interpreted. %
$\alpha_1^{(k)}$ corresponds to the maximum (normalised) power in the prior, and $\beta_k$ determines the rate of convergence (of the asymptote) for priors with sigmoidal mean functions. %
As expected, for the ideal curve ($k=1$) the mean of the prior tends to $\alpha_1^{(1)} = 1.021$. %
For $k=2$ the asymptote tends to $\alpha_1^{(2)} = 0.4623$; %
this provides a more accurate estimate of the maximum curtailed output (46.23\% rather than $\approx 50\%$). %
As expected, the rate of convergence is greater for the curtailment data ($\beta_2 = 28.8$) and lower for the ideal data ($\beta_1 = 11.4$); this can be visualised in the plots of the prior mean functions (black lines) \Cref{fig:hetOMGP}.

\subsection{Validation: more turbines and curtailments}\label{s:more}

\begin{figure}[pt]
\centering
\includegraphics[width=\textwidth]{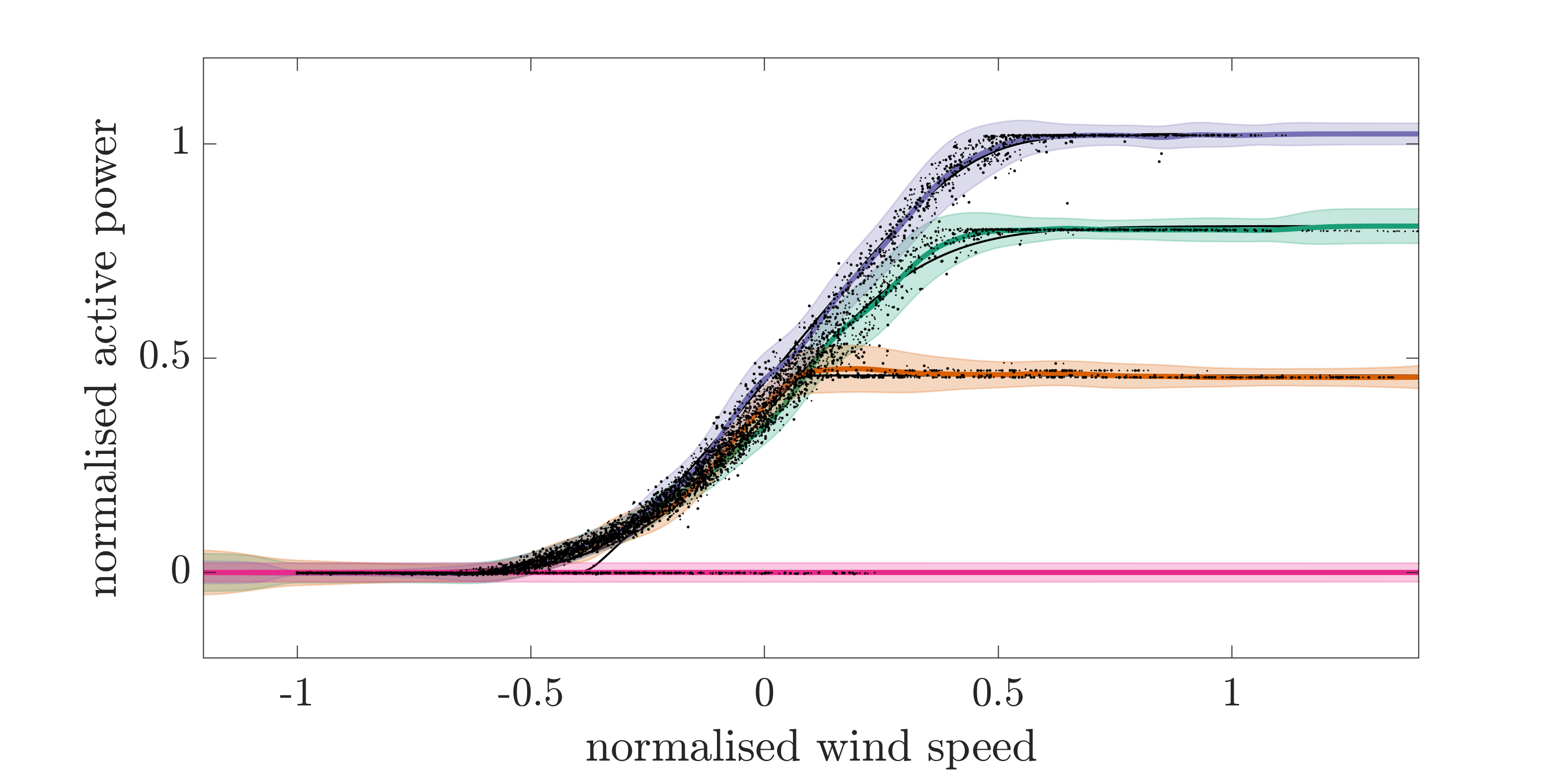}
\caption{Heteroscedastic OMGP of curtailed data from an alternative group of turbines, also exhibiting 80\% curtailment. Black lines show the mean functions of the prior $\bm{m}^{(k)}$. The green, orange, purple, and pink lines show the predictive mean $\bm{\mu}^{(k)}_*$, and shaded regions show three-sigma of the predictive variance $\diag(\bm{\Sigma}^{(k)}_*)$. Small {$\bm{\cdot}$} markers show the test set, and larger $\bullet$ markers show the training set.}\label{fig:80curt}
\end{figure}

To demonstrate the flexibility of the model it is used to infer $K > 3$ latent functions, associated with a \textit{separate} group of turbines in the wind farm. %
As before, the turbines exhibit normal,  50\% curtailment, and zero-power relationships; however, an 80\% curtailment is also observed. %
The priors of the OMGP are defined as in~\Cref{sec:OMGP}, with an additional soft-clip mean function component, such that $K=4$. The number of components is verified via cross-validation in \ref{a:cross-val}. %
A total of 9973 observations are sampled from the data, corresponding to a (selected) subset of four turbines over seven weeks. %
Approximately $1/3$ of the data are used for training and $2/3$ for testing. %
A representative model is learnt for the alternative latent functions, visualised in \Cref{fig:80curt}. %
The same metrics are presented in~\Cref{t:metrics80} to highlight improvements. %
Again, the hyperparameters of the OMGP are interpretable: in particular, for the new curtailment $\alpha_1^{(k)} = 0.81$, corresponding approximately to 80\%.

\begin{table}[]
\centering
\begin{tabular}{l|cc|cc}

 \hspace{1em} & \multicolumn{2}{c}{Conventional regression} & \multicolumn{2}{c}{Mixture of regressions} \\   
 \hspace{1em} & RVM & GP & OMGP & Het-OMGP \\
 \hline
 \hline


 NMSE & 10.46 & 10.19 & 0.15 & 0.15 \\
 MSD & 0.98 & 0.98 & 0.70 & 0.66
\end{tabular}\caption{Model performance metrics when $K=4$, including 80\% curtailed data.}\label{t:metrics80}
\end{table}

The validation experiments with four components ($K=4$) highlight that the OMGP can be used to represent a variety of curtailment relationships, supporting modelling and monitoring procedures for a wide range of data that should be expected in practice. %

\subsection{Towards population-based monitoring: entropy measures}

Considering applications of monitoring \textit{in situ}, the OMGP can be used to inform novelty detection and classification across the wind farm. %
Novel observations of wind speed and power (from the full 125 week monitoring period) can be compared to the OMGP. %
This approach to performance monitoring is an approach in population-based SHM, whereby a general model, referred to as the \textit{form} \cite{PBSHMMSSP1}, is used to represent the behaviour of members within a population. %
In this case, the form is the OMGP and the population is the wind farm.

When monitoring via the power curve, the error given the predicted output (e.g.\ \cref{eq:nmse,eq:msd}) can be used for novelty detection, as in \cite{PBSHMMSSP1,Papatheou2015,papatheou2017performance}. %
Alternatively, with the OMGP, given wind speed and power observations, measurements can be classified using $\hat{k}_*$ (\ref{eq:classify}). %
Additionally, the distribution $p(k_*\mid\bm{x}_*,\bm{y}_*,\mathcal{D}) = P(k_*\mid\bm{x}_*,\bm{y}_*,\mathcal{D})$ (\ref{eq:bayes_rules_classify}) is informative from a monitoring perspective; %
this is the probability that $\{\bm{x}_*,\bm{y}_*\}$ were generated by component $f^{(k)}$ in the mixture. %
In other words, the probability that new data correspond to: 

\begin{itemize} 
\item the normal curve $(k_*=1)$, 
\item 50~\% curtailment $(k_*=2)$, 
\item or zero-power $(k_*=3)$. %
\end{itemize}

\begin{figure}[pt]
\centering
\includegraphics[width=.6\textwidth]{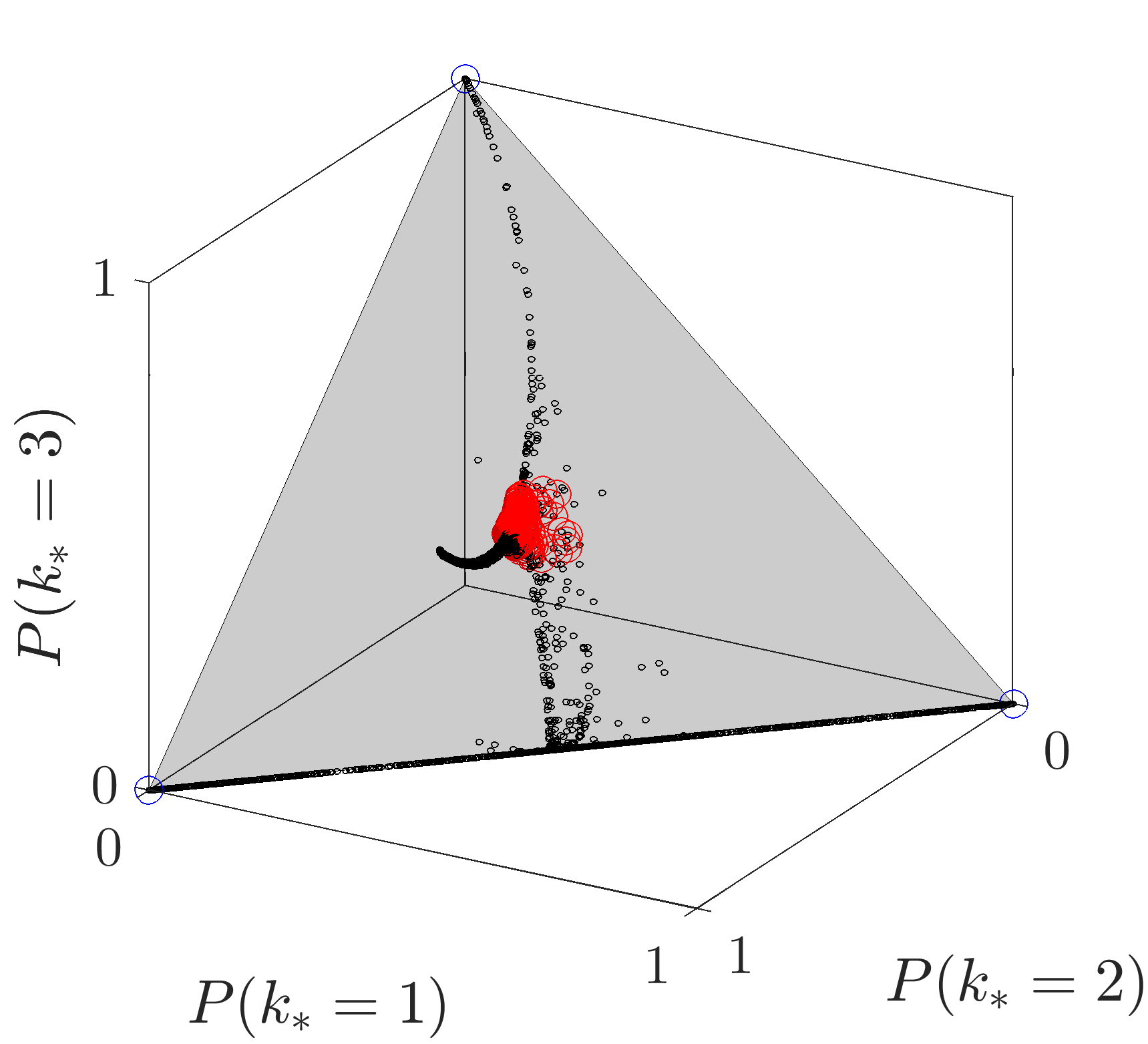}
\caption{The simplex (grey triangle) associated with the distribution $P(k_*\mid\bm{x}_*,\bm{y}_*,\mathcal{D})$. Points on the simplex represents observations of wind speed and power. Blue $\circ$ markers highlight low entropy points, red $\circ$ markers highlight high entropy points.}\label{fig:simplex}
\end{figure}

\noindent As $k_* \in \{1,2,3\}$, and $\sum_{k_*}^{K} P(k_*\mid\bm{x}_*,\bm{y}_*,\mathcal{D}) = 1 \quad \forall \{\bm{x}_*,\bm{y}_*\}$, it is possible to view power curve data as points on a 3D \text{simplex}, associated with the multinomial distribution $p(k_*\mid\bm{x}_*,\bm{y}_*,\mathcal{D})$. %
The grey triangle in \Cref{fig:simplex} visualises the simplex where points are observations from the test set (concerning the 50\% curtailment data). %
Although initially abstract, the plot is insightful from a monitoring perspective. %
It indicates that classes one and two (ideal and curtailed trends) are regularly confused, while class three (zero power) is equally confused with the others. %
This makes sense when inspecting \Cref{fig:hetOMGP}: the ideal and curtailed trends are similar up to a normalised wind speed of zero, while the zero-power trend is indistinguishable from $k=1$ and $k=2$ at low wind speeds. %

Given this distribution, the Shannon-entropy can be used as a measure of \textit{uncertainty} to indicate if it is likely that new data were generated by latent functions within the OMGP,

\begin{align}
 H(k_*) = - \sum^K_{j=1} P(k_*=j\mid\bm{x}_*,\bm{y}_*,\mathcal{D}) \log P(k_*=j\mid\bm{x}_*,\bm{y}_*,\mathcal{D})
\end{align}

\noindent With regard to the simplex in \Cref{fig:simplex}, each corner of the triangle relates to \textit{low} entropy, corresponding to data that are classified  with \textit{certainty} (as $k_*=1$, $k_*=2$, or $k_*=3$). On the other hand, the centre corresponds to \textit{high} entropy, i.e.\ observations for which each component is equally likely (or none at all). %
During monitoring, high entropy data can be investigated, as it is unclear which component generated them. %
Examples of high and low entropy data given the test set are shown by red and blue markers respectively in \Cref{fig:simplex}. %
Following investigation, if it appears that new data correspond to an additional latent function (not yet included in the \textit{form} of the wind farm) the mixture can be updated accordingly by adding a component, such that $K \leftarrow K+1$. %
Ideas behind modelling and updating the \textit{form} for a wind farm population (and subsequent monitoring) are the focus of future work.

\section{Conclusions}\label{s:conc}

A novel data-driven model for wind turbine power data has been proposed. %
Critically, the method is capable of representing wind/power measurements including both \textit{curtailed} and ideal operation. %
This is an alternative to the conventional approach, which filters out (and removes) the curtailed (SCADA) data. %
Consequently, the model should be representative of \textit{in situ} behaviour, rather than ideal operation only. %

A mixture of Gaussian processes infers \textit{multivalued} wind-power relationships without labels to associate data to functions. %
Each function corresponds to a different operational condition (power curve) for a wind farm population. %
The algorithm is unsupervised, as labels to define which trend (ideal, curtailed, etc.) generated each each of the measurements are not required; this information was not available in the experiments here. %
For each function in the mixture, input dependent noise is considered, a critical consideration when modelling power curve data. %
The model is applied to measurements from an operational wind farm, and it is shown to generalise well, representing future measurements from the population for various sets of turbines and curtailments. %
Finally, ideas for population-based power curve monitoring procedures (considering entropy measures) are introduced and discussed. %

\section*{Acknowledgements}
The authors gratefully acknowledge the support of the UK Engineering and Physical Sciences Research Council (EPSRC) through Grant references EP/R003645/1, EP/R004900/1, EP/R006768/1.

\bibliographystyle{unsrtnatemph}
\bibliography{ref}

\appendix

\section{Type-II Maximum Likelihood}\label{a:typeII}
Gaussian process hyperparameters, $\bm{\theta} = \left\{\beta, \bm{\alpha}, \sigma_f, l, \sigma \right\}$, are (typically) optimised through empirical Bayes. %
This involves maximising the marginal likelihood of the model,

\begin{align}
p(\bm{y}\mid\bm{x}\;;\;\bm{\theta}) &= \int{p(\bm{y}\mid\bm{x},\bm{f})p(\bm{f}\mid\bm{x})\;d\bm{f}} \nonumber\\
&= \mathcal{N}\left( \bm{y} \;;\; \bm{m},\;\bm{K_{xx}} + \bm{R}\right) \label{eq:ml}
\end{align}

\noindent By marginalising (integrating) out the latent function values $\bm{f}$, this moves a level up the Bayesian hierarchy -- mitigating issues of overtraining through the Bayesian Occam's razor \cite{rasmussen2001occam}. %
An optimisation of this objective should lead to a minimally-complex model given the observed training data; %
for convenience and numerical stability, this is implemented as a minimisation of the negative-log-marginal-likelihood w.r.t\ $\bm{\theta}$,

\begin{align}
\bm{\hat{\theta}} &= \argmin_{\bm{\theta}} \left\{ - \log{p\left(\bm{y}\mid\bm{x}\,;\,\bm\theta\right)} \right\} \\[1em]
- \log p\left(\bm{y}\mid\bm{x}\,;\,\bm\theta\right) &\triangleq - \log{\mathcal{N}\left( \bm{y} \mid \bm{m},\;\bm{K_{xx}} + \bm{R}\right)} \nonumber\\
&=  \frac{1}{2} \left(\bm{y} - \bm{m}\right)^{\top}\left(\bm{K_{xx}} + \bm{R}\right)^{-1}\left(\bm{y} - \bm{m}\right) \ldots \nonumber\\
& \qquad\qquad\qquad + \frac{1}{2}\log \left|\bm{K_{xx}} + \bm{R}\right| + \frac{N}{2} \log2\pi\label{eq:lml}
\end{align}

\noindent The terms in \cref{eq:lml} have an interpretable meaning: the first is a data \textit{fit} (or error) term, the second is a model complexity term, and the last is a constant \cite{murphy2012machine}. %

\section{Noise-Process Optimisation}\label{a:optim-HetGP}
\noindent Summarised from \cite{kersting2007most}:
\begin{enumerate}
    \item given $\mathcal{D}$, learn an initial homoscedastic GP: $G_1$ with hyperparameters $\mathcal{\bm{\theta}}$
    \item given $G_1$, estimate the empirical (log) input-dependent noise $\bm{g}^\prime$ at the inputs $\bm{x}$ using \cref{eq:g_est}
    \item given $\left\{\bm{g}^\prime, \bm{x}\right\}$, learn the noise process: $G_2$ with hyperparamters $\bm{\zeta}$
    \item given $\mathcal{D}$, estimate the heteroscedastic GP $G_3$, using $G_2$ to define $r(x_i)$ for the noise kernel
    \item if not converged, set $G_1 \leftarrow G_3$, and go to step 2. 
\end{enumerate}

\section{RVM Benchmark}\label{a:benchmarks}

The Relevance Vector Machine (RVM) follows the implementation of \citet{tipping2001sparse}, with a radial-basis function kernel, %
$$
k(x_i, x_j) = \exp{\left\{-\gamma(x_i - x_j)^2\right\}}
$$
The hyperparameter $\gamma$ of the kernel is determined by 5-fold cross-validation~\cite{murphy2012machine}. %
In the first (50\% curtailed) experiments, an optimal value was $\gamma=2$, while in the second experiments (80\% curtailed) $\gamma = 1.6$. %

\section{Example Cross-Validation}\label{a:cross-val}
\begin{figure}[pt]
\centering
\includegraphics[width=.6\textwidth]{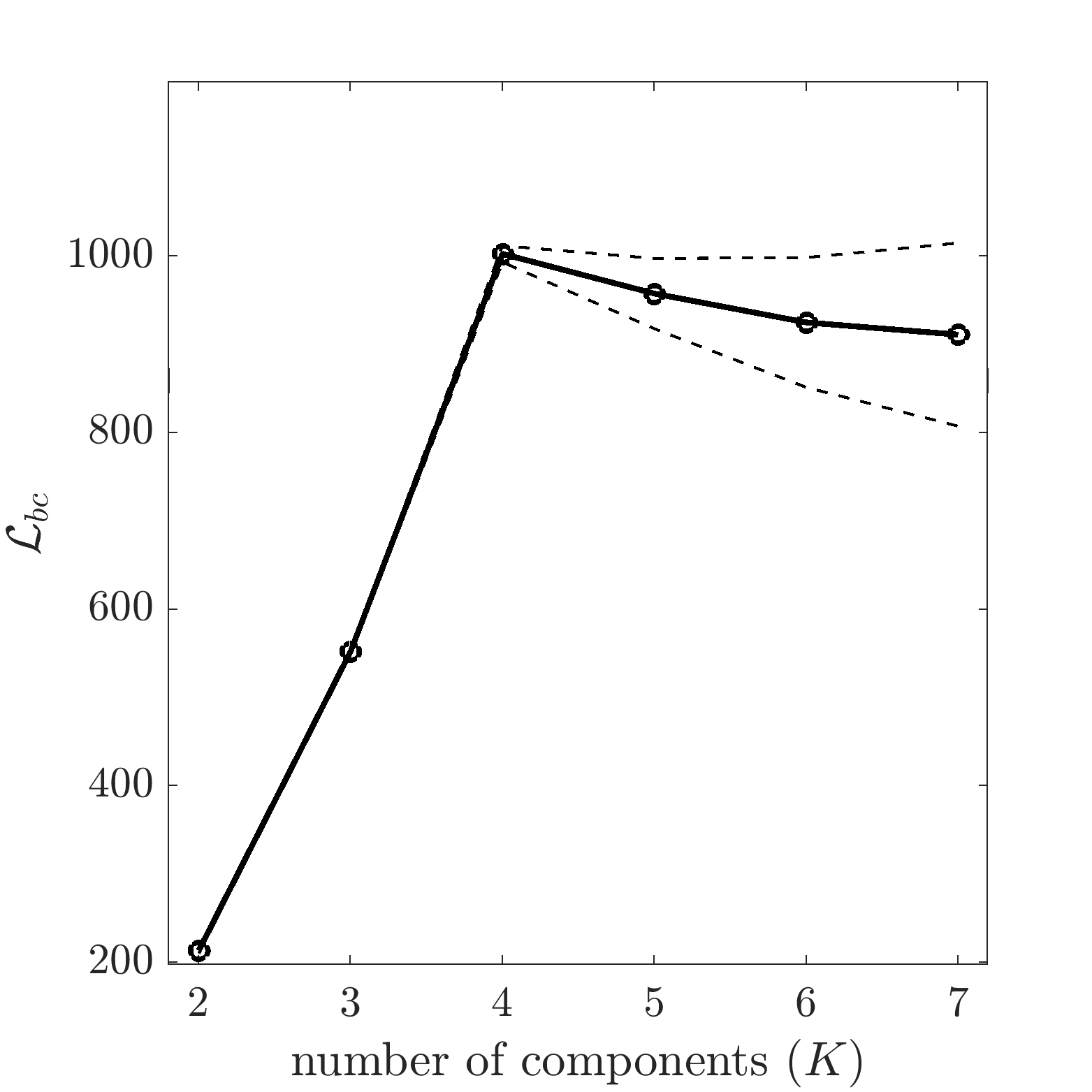}
\caption{Monitoring the corrected lower bound $(\mathcal{L}_{bc})$ on the marginal likelihood while the number of components increases. The solid line represents the mean and the dashed line shows three-sigma standard deviation (12 repeats).}\label{fig:cross-val}
\end{figure}

The corrected lower bound (\ref{eq:Lvb}) can be monitored to indicate an appropriate number of components for the OMGP. %
This is shown in \Cref{fig:cross-val} for the model in \Cref{s:more}. The maximum value of the lower bound corresponds to a mixture with four components ($K=4$) -- in correspondence with the prior intuition. %

\end{document}